\documentclass[lettersize,journal]{IEEEtran}
\usepackage{amsmath,amsfonts}
\usepackage{algorithmic}
\usepackage{algorithm}
\usepackage{array}
\usepackage[caption=false,font=normalsize,labelfont=sf,textfont=sf]{subfig}
\usepackage{textcomp}
\usepackage{stfloats}
\usepackage{url}
\usepackage{verbatim}
\usepackage{graphicx}
\usepackage{cite}
\usepackage[table, dvipsnames]{xcolor}
\newcommand{\red}[1]{{\color{red}#1}}

\definecolor{Gray}{gray}{0.85}
\definecolor{mylightgreen}{RGB}{198, 223, 200}
\definecolor{lightblue}{RGB}{230, 231, 253}
\definecolor{mygray}{gray}{.9}
\definecolor{ggray}{RGB}{127,127,127}
\definecolor{reda}{RGB}{192,0,0}
\definecolor{redb}{RGB}{217,148,143}
\definecolor{myyellow}{RGB}{190,144,0}
\definecolor{mygreen}{RGB}{80,100,40}
\definecolor{myblue}{RGB}{30,90,100}
\definecolor{apurple}{RGB}{102, 77, 166}
\definecolor{agreen}{RGB}{117, 175, 92}
\definecolor{ablue}{RGB}{66, 115, 147}
\definecolor{ablue}{RGB}{66, 115, 147}
\definecolor{ablack}{RGB}{57, 57, 57}
\definecolor{agray}{RGB}{101, 101, 101}
\definecolor{LightCyan}{HTML}{91bee1}
\definecolor{key_words}{HTML}{ec008d}
\definecolor{annotation}{HTML}{009900}

\usepackage{color}
\usepackage{pifont}       
\usepackage{bbding}       
\usepackage{fontawesome}

\usepackage{algorithm}
\usepackage{algorithmic}

\usepackage{newfloat}
\usepackage{listings}
\usepackage[colorlinks,bookmarksopen,bookmarksnumbered,citecolor=green, linkcolor=red, urlcolor=blue]{hyperref}
\usepackage{cleveref}
\crefname{figure}{Fig.}{Fig.}
\crefname{table}{Tab.}{Tab.}
\crefname{section}{Sec.}{Sec.}
\crefname{algorithm}{Alg.}{Alg.}
\usepackage{booktabs}
\usepackage{pifont}
\usepackage{multirow}
\usepackage{times}

\begin{document}

\title{Content-driven Magnitude-Derivative Spectrum Complementary Learning  for Hyperspectral \\ Image Classification}

\author{Huiyan Bai, Tingfa Xu$^{*}$, Huan Chen, Peifu Liu, Jianan Li$^{*}$

\thanks{Huiyan Bai, Tingfa Xu, Huan Chen, Peifu Liu, and Jianan Li are with the Beijing Institute of Technology, Beijing 100081, China, and also with the Key Laboratory of Photoelectronic Imaging Technology and System, Ministry of Education of China, Beijing 100081, China. (Email: baihuiyan0821@163.com, ciom\_xtf1@bit.edu.cn, 3220235096@bit.edu.cn, bitlpf@163.com, lijianan@bit.edu.cn).

Tingfa Xu is also with Big Data and Artificial Intelligence Laboratory, Beijing Institute of Technology Chongqing Innovation Center, Chongqing 401135, China.}
\thanks{$^{*}$Corresponding authors: Tingfa Xu and Jianan Li.}}

\markboth{Journal of \LaTeX\ Class Files,~Vol.~14, No.~8, August~2021}%
{Shell \MakeLowercase{\textit{et al.}}: A Sample Article Using IEEEtran.cls for IEEE Journals}


\maketitle

\begin{abstract}
Extracting discriminative information from complex spectral details in hyperspectral image (HSI) for HSI classification is pivotal. While current prevailing methods rely on spectral magnitude features, they could cause confusion in certain classes, resulting in misclassification and decreased accuracy. We find that the derivative spectrum proves more adept at capturing concealed information, thereby offering a distinct advantage in separating these confusion classes. Leveraging the complementarity between spectral magnitude and derivative features, we propose a Content-driven Spectrum Complementary Network based on Magnitude-Derivative Dual Encoder, employing these two features as combined inputs. To fully utilize their complementary information, we raise a Content-adaptive Point-wise Fusion Module, enabling adaptive fusion of dual-encoder features in a point-wise selective manner, contingent upon feature representation. To preserve a rich source of complementary information while extracting more distinguishable features, we introduce a Hybrid Disparity-enhancing Loss that enhances the differential expression of the features from the two branches and increases the inter-class distance. As a result, our method achieves state-of-the-art results on the extensive WHU-OHS dataset and eight other benchmark datasets.
\end{abstract}


\section{Introduction}
\IEEEPARstart{A} hyperspectral image contains hundreds of near-continuous spectral bands, which can provide a wealth of information. 
The rich information inherent in HSI facilitates its widespread application across various fields~\cite{qin2024dmssn, liu2023spectrum, chen2023spectral, zhou2023rdfnet, wang2024opto}, especially land cover classification and urban planning~\cite{vivone2023multispectral}. The presence of spectral redundancy and overlap, combined with similar reflectance distributions in specific spectral bands, greatly diminishes the spectral utility, rendering HSI classification highly challenging.

\begin{figure}[t]
  \centering
  \includegraphics[width=1.0\linewidth]{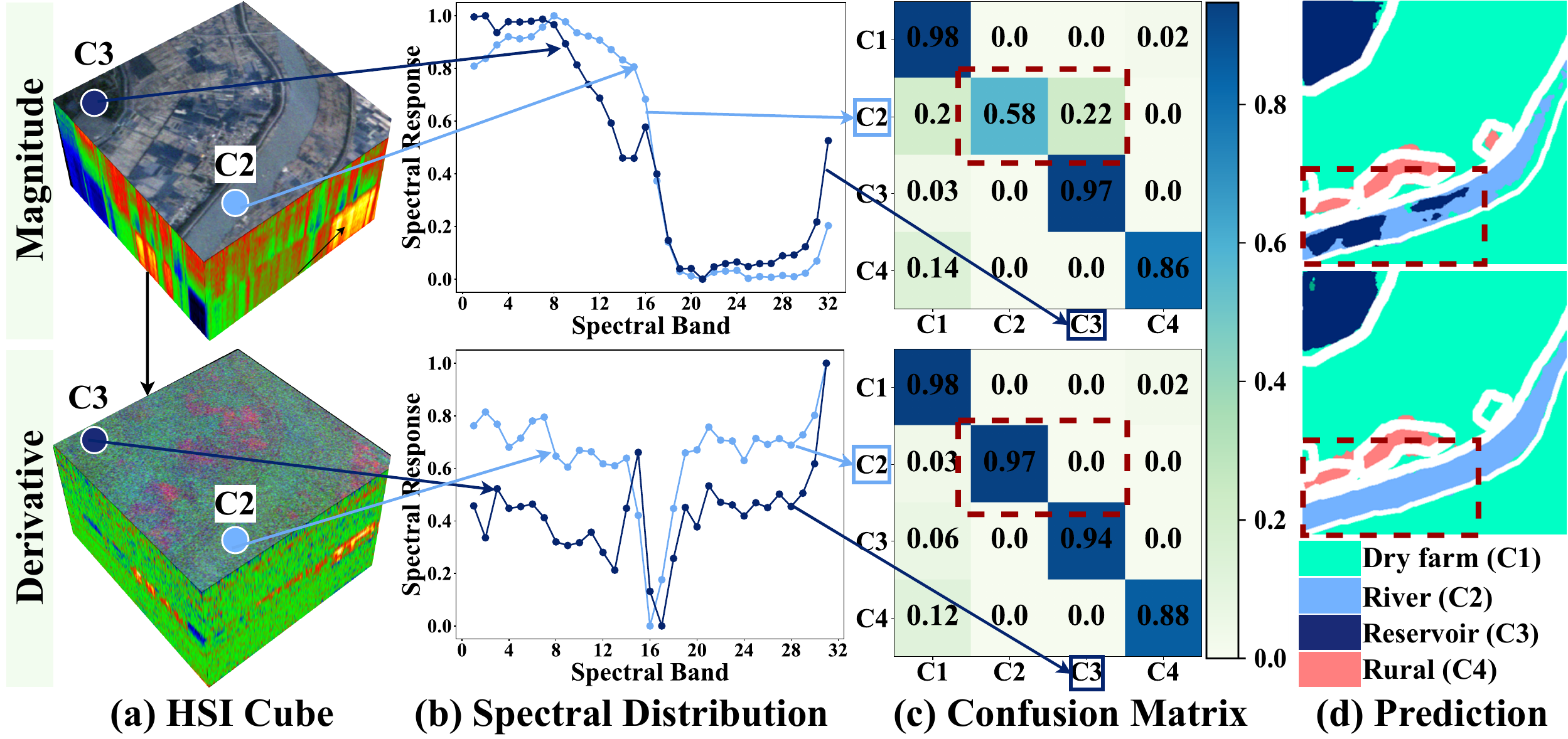}
  \caption{\textbf{(a)} HSI cubes present the visualization of spectral magnitude and derivative features. In \textbf{(b)}, spectral distributions of C2 and C3 are compared, showing strong similarity in magnitude distribution but significant differences in derivative. Subsequently, the classification confusion and accuracy decline in \textbf{(c)} of magnitude reflects its insufficient feature separation ability, where the accurate classification in derivative precisely compensates for it. Additionally, the visualization of the classification results \textbf{(d)} also confirms this characteristic.}
  \label{Fig: Motivation}
\end{figure}

The evolution of deep learning has sparked considerable interest in CNN-based techniques for HSI classification. While 1D-CNN excels at extracting spectral information through one-dimensional convolution in the spectral domain, it may not fully capture the spatial contextual information. 2D-CNN, on the other hand, is adept at extracting local spatial information around target pixels, thus enhancing spatial feature utilization. However, it often necessitates dimensionality reduction, which can disrupt spectral structures and lead to information loss.
To address these limitations, recent methods incorporate advanced techniques such as spatial and spectral attention mechanisms, multi-scale features, and context-aware modules~\cite{xu2018spectral,wang2020adaptive}, all aimed at refining the extraction of both local and global features by effectively leveraging spectral and contextual information. 
Due to spectral redundancy, some spectral bands may exhibit similar reflectance intensity distributions. For neural networks, extracting effective spectral structures from these complex spectral data can be challenging, potentially leading to insufficient utilization of spectral complexity. Consequently, utilizing a single spectral input in this manner may result in feature confusion for certain classes. \cref{Fig: Motivation} displays the spectral magnitude and derivative feature distributions, along with a comparison of the classification confusion matrix and the visualization results for the given scenario. Illustrated in the top row of \cref{Fig: Motivation}, similar spectral distributions in spectral magnitude features cause confusion between several classes, causing misclassification and reduced accuracy.

To address the issue of underutilized spectral information, some studies have developed spectral attention networks. These networks~\cite{liu2022central, xu2023multiscale} leverage attention mechanisms to capture spectral features and reweight them to suppress redundant spectral information, enhancing the extraction of discriminative features. Although attention mechanisms can effectively capture discriminative features, they still fall short in handling spectral confusion.  We propose that the incorporation of additional spectral complement information can establish a comprehensive spectral perception and fully leverage spectral data, thereby further assisting in the learning of spectral features.
Through in-depth spectral analysis, spectral derivative features are instrumental in addressing overlapping spectral characteristics and enhancing spectral contrast. Concurrently, spectral derivatives reveal differences between adjacent spectra, which can amplify minor spectral features, extract hidden information, and more effectively detect spectral nuances. As illustrated in \cref{Fig: Motivation}, the outstanding performance of derivative features in discerning features in certain spectral magnitude-confused instances implies their complementarity, thereby enhancing the discrimination of features that are confounded under a single spectrum. Building upon this, we desire to implement fusion between these features to extract more favorable discriminative information.

Given the considerations outlined above, we propose a novel Content-driven Spectrum Complementary Network, dubbed CSCN, whose fundamental component is a Magnitude-Derivative Dual Encoder with identical structures. We employ spectral magnitude and derivative features as inputs, which show advantages in distinct classes. The encoder is responsible for the joint learning of magnitude and derivative spectrum to extract rich complementary information.

As mentioned earlier, the magnitude features exhibit prominent advantages in certain regions while the derivative features show significant advantages in others. To better utilize the complementary information, we devise a novel Content-adaptive Point-wise Fusion Module capable of achieving adaptive fusion at a point-wise for the dual encoder. At each point, features from both branches are adaptively fused based on their content. More specifically, the feature of each point on the fused feature map from the previous stage is defined as a \textit{query}, and the features from the same point on both branches are regarded as \textit{keys}. By modeling fine-grained interactions and long-range dependencies between them, the \textit{query} feature is continuously updated and refined as a weighted sum of the \textit{keys} features. Consequently, the content at each point independently determines the selected feature sources, enabling the generation of a dense content-aware feature map that efficiently leverages the utilization of complementary information. 

Furthermore, leveraging the complementary relationship demonstrated by the features extracted from the dual-encoder, which offers rich discriminative information, our objective is to enhance diversity in feature representation. We investigate the correlation between the independent coarse predictions, dynamically selecting point-wise greater-contributing features to facilitate the supervised learning process. Additionally, leveraging class-specific feature representations, we incorporate class-wise contrastive loss during training, designed to understand the associations between features belonging to the same class, encouraging increased inter-class separation.

As a result, our method facilitates a comprehensive fusion of spectral magnitude and derivative features, effectively addressing class confusion from spectral overlap and similar distributions, and enhancing classification accuracy.
The efficacy of our method is proven through rigorous experiments on the challenging WHU-OHS dataset~\cite{WHU-OHS} and eight benchmarks, including IP, PU, WHU-Hi-HanChuan, WHU-Hi-LongKou~\cite{Zhong2020WHU-Hi}, HyRANK-Dioni, HyRANK-Loukia~\cite{karantzalos2018hyrank}, Augsburg and Berlin datasets~\cite{hong2023cross}. Outperforming existing methods, it excels at leveraging rich spectral information while capturing spatial details and effectively extracting discriminative features.

Our contributions can be summarized as follows:
\begin{itemize}
        \item  We propose a novel Dual-Encoder integrating spectral magnitude and derivative features to extract rich complementary information.

        \item We raise a Content-aware Point-wise Fusion Module, utilizing the complementary features from both branches adaptively based on point content. 
   
        \item We introduce a Hybrid Disparity-enhancing Loss to promote the exploration of disparities and preserve the distinctiveness of both features.
\end{itemize}

\section{Related Work}
\label{sec:related}

\subsection{Spectral Derivative}
In spectroscopy, spectral derivatives are crucial for detecting spectral variations. In the realm of HSI classification, they capture fine spectral details while reducing the effects of lighting and atmospheric conditions. Tsai \textit{et al.}~\cite{tsai1997derivative} emphasized the significant potential of derivative analysis in enhancing HSI processing. Bao \textit{et al.}~\cite{bao2013spectral} combined spectral magnitude with derivative features, and Kalluri \textit{et al.}~\cite{kalluri2010decision} fused these at the decision level to improve classification accuracy. Traditional methods combined spectral derivatives with classifiers such as Support Vector Machine and Multinomial Logistic Regression to extract spectral and spatial features, yet they were constrained by manual feature crafting and underutilized spectral details. In contrast, our approach integrates spectral derivatives with CNNs to deeply and efficiently mine spectral intricacies and spatial structures.

\subsection{CNN-based HSI Classification}
Existing HSI classification approaches can be divided into two main categories: traditional machine-learning methods and deep-learning methods. In traditional machine-learning methods, researchers typically extract features from hyperspectral data and then classify them using various classifiers. These methods can be further subdivided into those based on spectral features, such as Random Forest~\cite{ham2005investigation} and Support Vector Machines~\cite{melgani2004classification} and methods based on the joint features of spectral and spatial information, such as Superpixel~\cite{fang2015classification}, and Gray Level Co-occurrence Matrix~\cite{pesaresi2008robust}.
Given the high dimensionality of HSI, which can negatively impact classification performance, various dimensionality reduction algorithms have been proposed. These include Principal Component Analysis~\cite{prasad2008limitations}, and Linear Discriminant Analysis~\cite{ye2016l1}, which aim to minimize redundant information while retaining useful information.
Furthermore, due to the excessively high dimensionality of HSI, which can significantly impair the enhancement of classification results, researchers have proposed methods such as Principal Component Analysis~\cite{prasad2008limitations} and Linear Discriminant Analysis~\cite{ye2016l1}.

Deep neural networks autonomously distill robust, discriminative features from raw data. This has shifted HSI classification from traditional machine learning to deep learning paradigms. Chen \textit{et al.}~\cite{chen2014deep} pioneered deep learning in HSI, introducing a spectral-spatial classification method using Stacked Auto-encoders. Chen \textit{et al.}~\cite{chen2015spectral} later integrated Deep Belief Networks for feature extraction, advancing the field.
Following these innovations, diverse deep learning architectures—Convolutional Neural Networks, Capsule Networks, Generative Adversarial Networks, Recurrent Neural Networks, and Transformer—have been effectively applied to HSI classification.
Additionally, with the significant advancements in foundational models, SpectralGPT~\cite{hong2024spectralgpt}, designed specifically for spectral remote sensing, offers a transformative solution for the effective and comprehensive utilization of spectral data.

Among these, CNNs are widely recognized as essential frameworks for extracting discerning features from HSI, capturing both local and global characteristics effectively.
The application of 1D-CNNs in HSI classification is well-documented, while the prevalence of 2D-CNNs and 3D-CNNs for spatial and spectral feature extraction is on the rise. Researchers continuously refine these models to improve accuracy and efficiency~\cite{zhong2017spectral, li2020classification, roy2020attention}. 

The aforementioned models typically adopt patch-based learning frameworks, wherein patches generated by neighboring pixels overlap, thereby augmenting computational complexity and impeding learning speed. In contrast, our approach embraces a patch-free learning framework that utilizes the entire image as input, which aims to conduct feature extraction in an end-to-end, pixel-to-pixel manner, effectively reducing computational complexity.
SSFCN~\cite{xu2019beyond} and FreeNet~\cite{zheng2020fpga} have devised end-to-end HSI classification networks to harness global information. Previous works~\cite{shen2020efficient, zhu2021spectral, tu2024multi} utilize a long-short-term spectral capture module, non-local feature learning module, and multi-scale feature utilization module to better capture spectral and spatial information.

Despite the progress made, there is still scope for enhancing the comprehensive perception of rich spectral information. Our goal is to effectively extract discriminative information from complex spectral data to mitigate the classification errors induced by spectral confusion. Our method acknowledges the unique discriminative advantages of spectral derivative features among certain categories, which can excellently complement spectral magnitude features. The synergistic interaction of these two sets of features enables a more holistic spectral perception and feature extraction.

\subsection{Attention Mechanisms for HSI Classification}
The inception of attention mechanisms draws inspiration from the human brain's remarkable ability to selectively focus on and recognize signals that are most pertinent to a given task. Initially applied in the domain of natural language processing, attention mechanisms have now found extensive utilization in computer vision, enabling the capture of long-range contextual relationships and enhancing the network's capacity to represent global information effectively.

The widespread integration of self-attention mechanisms in computer vision has given rise to a novel class of non-local networks. This evolution began with Non-local neural networks~\cite{wang2018non} and has since been followed by \cite{xue2019danet, zhou2022canet}. These contributions have significantly fortified the network's ability to comprehend the global context of the scene, underscoring the immense potential of attention mechanisms in the realm of computer vision.

In the context of HSI classification, previous works~\cite{hang2020hyperspectral, liu2022central} utilize joint spectral-spatial attention modules, global-local attention mechanisms, and cross-layer stacked attention mechanisms to aid in the extraction of spectral-spatial features. In comparison with these, in our method, the features are fused in adaptive proportions, depending on their corresponding content, which enables better utilization of complementary information from the magnitude and derivative features, aiding in discriminative feature extraction.

\begin{figure*}[ht]
    \centering
    \includegraphics[width=0.95\linewidth]{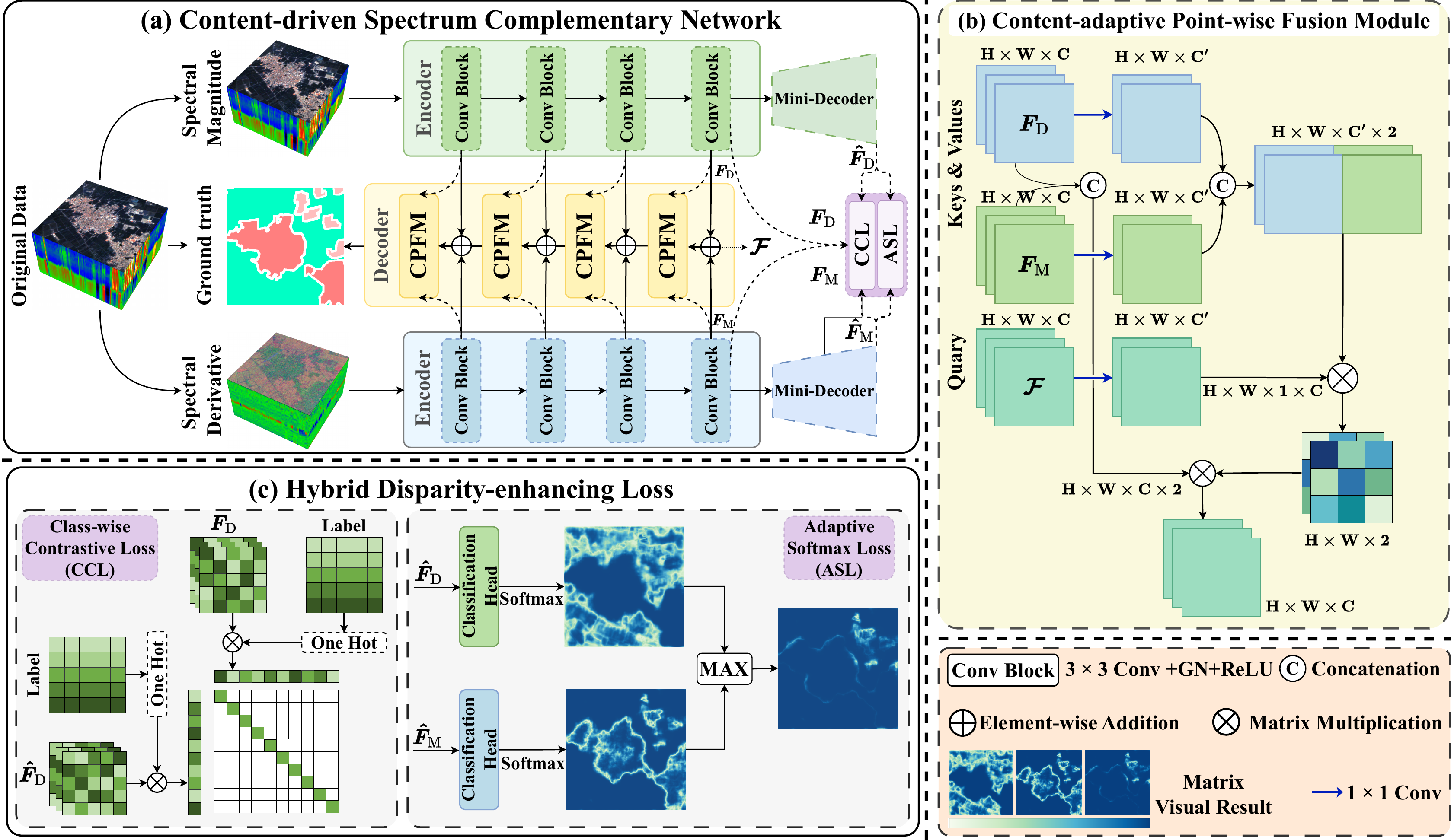}
        \caption{\textbf{(a)} Overall architecture of the proposed Content-driven Spectrum Complementary Network consists of the Magnitude-Derivative Dual Encoder, Content-adaptive Point-wise Fusion Module, and Hybrid Disparity-enhancing Loss. \textbf{(b)} The proposed fusion module receives features from the dual-encoder along with the fused features from the previous stage, refining and feeding them into the next stage through aggregation and refinement processes. \textbf{(c)} The proposed loss function integrates the rough predictions from the dual-encoder and receives features from the encoder and mini-decoder. This integration provides additional supervision to aid network learning. (In matrix visual results, darker colors indicate higher confidence in the classification results.)
}
    \label{Fig: Framework}
\end{figure*}

\section{Method}
Given original HSI data denoted by $\boldsymbol{I} \in \mathbb{R}^{\mathrm{H} \times \mathrm{W} \times \mathrm{B}}$, the objective of HSI classification is to employ the mapping function $\boldsymbol{\Phi}(\cdot)$
to generate a classification map represented as $\boldsymbol{T} \in \mathbb{R}^{\mathrm{H} \times \mathrm{W}}$:
\begin{equation}
    \boldsymbol{T} = \boldsymbol{\Phi}(\boldsymbol{I}),
\end{equation}
where $ {\mathrm{H} \times \mathrm{W}} $ represents spatial dimensions and $\mathrm{B}$ indicates the band number. 

We present a novel Content-driven Spectrum Complementary Network, dubbed CSCN, based on the Magnitude-Derivative Dual Encoder for HSI classification. The architecture of the network is depicted in \cref{Fig: Framework}. The dual encoder structure equips the network with spectral magnitude and derivative features. The Point-wise Content-adaptive Fusion Module facilitates the complete utilization of complementary information from both features, precisely selecting valuable discriminative information at each point from the dual encoder. The Hybrid Disparity-enhancing Loss is crafted to explore the internal relationship between the two branches, to increase or maintain their distinctiveness while maximizing inter-class separation.

\subsection{Magnitude-Derivative Dual Encoder}
We have noted that features derived from spectral magnitude and obtained from spectral derivative possess complementary characteristics. Consequently, we introduce a dual-encoder featuring distinct inputs for spectral magnitude and derivative features as depicted in \cref{Fig: Framework}~\red{(a)}. It is noteworthy that the two encoders have consistent feature extraction structures, with non-shared parameters.

\noindent \textbf{Spectral Derivative.} 
\label{Sec: Derivaitve}
Since the HSI classification dataset's hyperspectral data is discrete, we can represent spectral features as: $\boldsymbol{I} = \left \{ \boldsymbol{Y}_i \right \}^{\mathrm{B}}_{i=1}$, while $\boldsymbol{Y}_i$ denotes a matrix containing reflectance values for these $ {\mathrm{H} \times \mathrm{W}} $ pixels at the $i$-th band. The first-order spectral derivative at the $i$-th band ($j>i$) can be defined as:
\begin{equation}
    \boldsymbol{Y}' \left ( i \right )=
    \frac{\boldsymbol{Y}_j  - \boldsymbol{Y}_i}{j-i}=
    \frac{\boldsymbol{Y}_j  - \boldsymbol{Y}_i}{\Delta n}.
\end{equation}
where $j = i + \Delta n$, $\Delta n \in \mathbb{Z}^{+}$ is the step length. Hence, the first-order spectral derivative at $\Delta n$ step length can be expressed as follows: $\boldsymbol{I'} = \left \{ \boldsymbol{Y'}_i \right \}^{\mathrm{B} - \Delta n}_{i=1} \in \mathbb{R}^{\mathrm{H} \times \mathrm{W} \times (\mathrm{B} - \Delta n)}$.

Similarly, the second-order spectral derivative can be obtained through iterative computation of the first-order derivative. The second-order spectral derivative at the $i$-th band ($k>j>i$ and $k = j + \Delta n = i + 2 \Delta n$) can be defined as:
\begin{equation}
\begin{split}
    \boldsymbol{Y''} \left ( i \right ) &=
    \frac{\boldsymbol{Y'}_j  - \boldsymbol{Y'}_i}{j-i} =
    \frac{\frac{\boldsymbol{Y}_k  - \boldsymbol{Y}_j}{\Delta n} - \frac{\boldsymbol{Y}_j  - \boldsymbol{Y}_i}{\Delta n}}
    {\Delta n} 
    \\
    &= \frac{1}{\Delta n ^ 2} \left [ \boldsymbol{Y}_k - 2\boldsymbol{Y}_j + \boldsymbol{Y}_i \right ].
\end{split}
\end{equation}
Therefore, the second-order spectral derivative at $\Delta n$ step length can be expressed as follows: $\boldsymbol{I''} = \left \{ \boldsymbol{Y''}_i \right \}^{\mathrm{B} - 2 \Delta n}_{i=1} \in \mathbb{R}^{\mathrm{H} \times \mathrm{W} \times (\mathrm{B} - 2 \Delta n)}$.

We employ a first-order spectral derivative with $\Delta n=1$ to build the spectral derivative feature matrix.

\noindent \textbf{Dual Encoder Framework.}
\label{encoder}
Our encoder can be characterized as an effective stacking of $\rm{N}$ convolution blocks. The feature $\boldsymbol{F}$ of the $l$-th block can be characterized as follows: 
\begin{equation}
    \boldsymbol{F}_{l} = \boldsymbol{\sigma}\left(
    \boldsymbol{GN}\left(      
    \left(
    \boldsymbol{\psi}
    \left(
    \boldsymbol{F}_{l-1}
    \right)
    \right)
    \right)
{}    \right),
\end{equation}
where $\boldsymbol{\psi}(\cdot)$, $\boldsymbol{GN}(\cdot)$, and $\boldsymbol{\sigma}(\cdot)$ denotes $3 \times 3$ convolution, group normalization and ReLU. Subsequently, we utilize a $3 \times 3$ convolution with a stride of 2, coupled with a ReLU, to perform downsampling on the features, which can be denoted as:
\begin{equation}
    \boldsymbol{F}_{l} = \boldsymbol{\sigma}\left(      
    \left(
    \boldsymbol{\rho}
    \left(
    \boldsymbol{F}_{l}
    \right)
    \right)
    \right),
\end{equation}
where $\boldsymbol{\rho} ( \cdot )$ denotes $3 \times 3$ convolution with a stride of 2. As the process of downsampling progresses, the features continue to deepen.

\begin{figure*}[ht]
  \centering
  \includegraphics[width=0.75\linewidth]{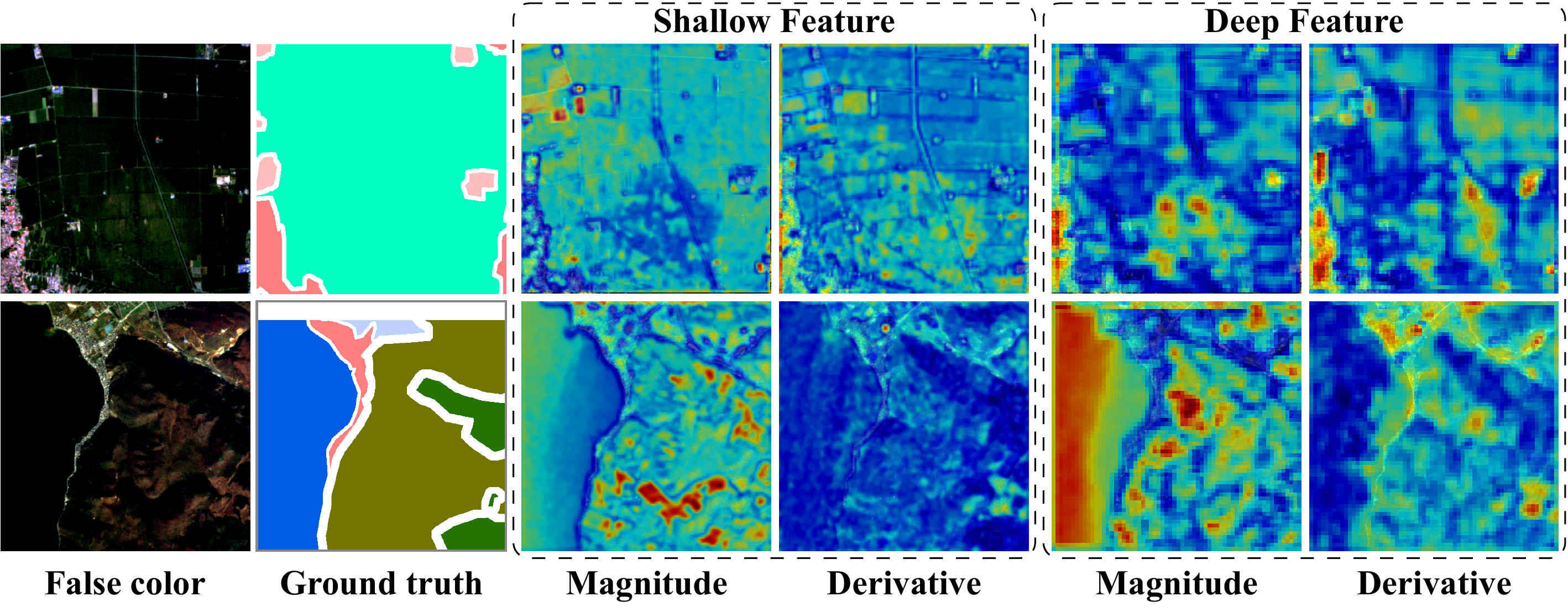}
  \caption{Comparison of \textbf{Feature Maps} from dual encoder as well as shallow and deep features. Deep features contain richer high-order semantic information, resulting in more explicit feature representation. It tends to be consistent in certain regions (first row) and exhibits notable differences in others (second row). To extract more complementary information, our goal is to enhance or preserve feature diversity between the two branches.}
  \label{Fig: FeatureMap}
\end{figure*}

\subsection{Content-adaptive Point-wise Fusion Module.}
\label{Fusion}
In order to fully leverage detailed spatial features and semantically rich deep features, we devise a progressive feature refinement pathway. This pathway involves stacking Content-adaptive Point-wise Fusion Module (CPFM), which adaptively bolsters the network's point-wise feature representation capabilities, and merges interactive details. The proposed fusion module integrates three features from both the dual-encoder and the decoder, amalgamating them into new enhanced features that are then forwarded to the next fusion module.

As can be seen in \cref{Fig: Framework}~\red{(b)}, our proposed fusion module, at each point, adaptively fuses feature from both branches based on their corresponding content through a novel point-wise attention mechanism, thereby achieving efficient utilization of complementary information. We further refine this module into Dual Feature Embedding, Content-adaptive Weight Generation, and Point-wise Feature Fusion.

\noindent\textbf{Dual Feature Embedding.}
\label{feature embedding}
Due to the presence of $\rm{N}$ convolutional blocks in the encoder, let's divide the feature fusion process into $\rm{N}$ stages. Taking the last stage as an example, we map the features from the dual encoder into a consistent feature dimension $\mathrm{C_f}$ across stages:
\begin{equation}
    \boldsymbol{X}_{\rm{M}} = \boldsymbol{\Psi}(\boldsymbol{F}_{\rm{M}}),
\end{equation}
where $\boldsymbol{\Psi} ( \cdot )$ denotes $1 \times 1$ convolution. The implementation of $\boldsymbol{X}_{\rm{D}}$ is similar to that of $\boldsymbol{X}_{\rm{M}} \in \mathbb{R}^{\mathrm{H} \times \mathrm{W} \times \mathrm{C_f}} $, where the corner symbols $\rm{M}$ and $\rm{D}$ denote magnitude and derivative.

\noindent\textbf{Content-adaptive Weight Generation.}
The purpose of our fusion module is to model long-range dependencies between the contents of each point from the dual-encoder and integrate the complementary information between them. 

The complementarity fusion refines the response at point $t$ in the fused features from the previous stage $\boldsymbol{\mathcal{F}} \in \mathbb{R}^{\mathrm{H} \times \mathrm{W} \times \mathrm{C_f}}$, that is the \textit{query} feature $\boldsymbol{\mathcal{F}}^{t}$, as a weighted sum of the responses at point $t$ in dual encoder feature representation, namely, \textit{key} feature $ \left \{ \boldsymbol{X}_{\rm{M}}^t, \boldsymbol{X}_{\rm{D}}^t \right \}$.

For the spectral magnitude features, the $\boldsymbol{X}_{\rm{M}}^t$ is projected onto the $\boldsymbol{\mathcal{K}}(\boldsymbol{X}_{\rm{M}}^t) = \boldsymbol{W}_{\rm{k}} \boldsymbol{X}_{\rm{M}}^t$, while the $\boldsymbol{\mathcal{F}}^{t}$ is projected onto the $\boldsymbol{\mathcal{Q}}(\boldsymbol{\mathcal{F}}^{t}) = \boldsymbol{W}_{\rm{q}} \boldsymbol{\mathcal{F}}^{t}$. The point-wise linear embedding $\boldsymbol{\mathcal{K}}$ and $\boldsymbol{\mathcal{Q}}$ is implemented by 1×1 convolution with the learnable weight matrix: $\boldsymbol{W}_{\rm{q}}$ and $\boldsymbol{W}_{\rm{k}}$. The weight $\boldsymbol{\mathcal{A}}^{t}_{\rm{M}}$ computes the similarity between them, which is implemented as follows:
\begin{equation}
    \boldsymbol{\mathcal{A}}^{t}_{\rm{M}} = \boldsymbol{\mathrm{Softmax}}{
    \left (
    \frac{{\boldsymbol{\mathcal{Q}}(\boldsymbol{\mathcal{F}}^{t})}^{\top} {\boldsymbol{\mathcal{K}}(\boldsymbol{X}_{\rm{M}}^t)}}
    {\sqrt{\mathrm{C_f}}}
    \right )}.
\end{equation}

\noindent\textbf{Point-wise Feature Fusion.}
With the same process, for the derivative features, $\boldsymbol{\mathcal{A}}^{t}_{\rm{D}}$ is implemented. Thus, the point fusion feature can be represented as:
\begin{equation}
    \boldsymbol{\hat{\mathcal{F}}}^{t} = \boldsymbol{\mathcal{A}}^{t}_{\rm{M}} \boldsymbol{X}_{\rm{M}}^t + \boldsymbol{\mathcal{A}}^{t}_{\rm{D}} \boldsymbol{X}_{\rm{D}}^t.
\end{equation}

Intuitively, this fusion recalibrates weights in dual-encoder feature fusion. The dynamic selection enables adaptive fusion based on each point's content, significantly enhancing dual-branch interaction modeling, leveraging complementary information, and facilitating fine-grained interaction between branches.

\subsection{Hybrid Disparity-enhancing Loss}
\label{Loss}
As demonstrated in \cref{Fig: Framework}~\red{(c)}, we incorporate a straightforward Mini-Decoder, consisting of a simple combination of $3 \times 3$ convolutional layers and upsampling layers, after each of the two Encoders. Leveraging its features and coarse predictions, we design a Hybrid Disparity-enhancing Loss. It can be characterized as an Adaptive Softmax Loss that increases the difference in two-branch feature expression and guarantees that both branches maintain a degree of independence and distinctiveness. Simultaneously considering the aggregation of information within the same class and the discrimination of information between different classes enhances the discriminative learning of spectral-spatial features, thereby amplifying inter-class differences.

\noindent \textbf{Adaptive Softmax Loss.}
\cref{Fig: FeatureMap} illustrates the comparative representations between two branches as well as shallow and deep features. Deep features contain richer high-order semantic information, resulting in more explicit feature representations. While certain regions exhibit a tendency towards consistent feature expressions, there are noticeable differences in feature representations in some areas. To provide the network with more complementary information, it is desirable to either increase or maintain the feature differences between the two branches.
We harness the coarse predictions from the two Mini-Decoders to implement the Adaptive Softmax Loss. With the two sets of coarse predictions represented as $\left \{{\boldsymbol{R}}_{\rm{M}}, {\boldsymbol{R}}_{\rm{D}} \right \} \in \mathbb{R}^{\mathrm{H} \times \mathrm{W} \times \mathrm{C_g}}$, independently undergoing Softmax calculations, we create a new adaptive matrix $S$ by selecting the maximum corresponding result between the two confidence matrices on a point-by-point basis. This approach encourages each branch to learn its advantageous features, thereby maintaining the distinctiveness and complementarity of feature representations. The matrix $S$ is then utilized in conjunction with the log and Negative Log-Likelihood (NLL) loss to assist the network in learning. With the label $G$, the loss can be described as:
\begin{equation}
    {\mathcal{L}}_{\rm{AS}}({\boldsymbol{S}}, {\boldsymbol{G}}) =
    -{\sum^{\rm{P}}_{p=1}}{\sum^{\rm{C_g}}_{c=1}}
    {\hat{\boldsymbol{G}}_{p,c}}(log({\boldsymbol{S}}_{p,c})),
\end{equation}
where, $\boldsymbol{S}_{p,c}$ represents the probability distribution of the pixel representing the $p$-th instance of class $c$:
\begin{equation}
    {\boldsymbol{S}}_{p,c}({\boldsymbol{R}}_{\rm{M}}, {\boldsymbol{R}}_{\rm{D}}) = {\boldsymbol{Max}} \left(
    \frac
    {{\boldsymbol{e}}^{{\boldsymbol{R}}_{\rm{M}}}_{p,c}}
    {\sum^{\mathrm{C_g}}_{c=1} {{\boldsymbol{e}}^{{\boldsymbol{R}}_{\rm{M}}}_{c}}},
    \frac
    {\boldsymbol{e}^{\boldsymbol{R}_{\rm{D}}}_{p,c}}
    {\sum^{\mathrm{C_g}}_{c=1} \boldsymbol{e}^{{\boldsymbol{R}}_{\rm{D}}}_{c}}
    \right).
\end{equation}
Here, $\mathrm{P}$ represents the number of pixels, $\mathrm{C_g}$ denotes the number of classes contained in the label, and $\hat{\boldsymbol{G}}$ represents the one-hot encoding of the label. To more clearly illustrate our loss design, the pseudocode is presented as shown in \cref{Alg: ASL}.

\begin{algorithm}[!t]
\centering
\renewcommand{\arraystretch}{0.5}
\caption{Pseudo-Code of Adaptive Softmax Loss.} 
\label{Alg: ASL}
\begin{lstlisting}[language={Python}]
# p_M: coarse prediction from Magnitude-Encoder
# p_D: coarse prediction from Derivative-Encoder
# l: label

def _adaptive_softmax_loss(p_M, p_D, l):
   p_M = Softmax(p_M, dim=1)
   p_D = Softmax(p_D, dim=1)
   p = Cat((p_M.unsqueeze(dim=-1), 
            p_D.unsqueeze(dim=-1), dim=-1)
   p = max(p, dim=-1, keepdim=False)
   p_log = log(p)
   loss = NLLLoss(p_log, l)
   return loss
\end{lstlisting}
\end{algorithm}

\begin{algorithm}[!t]
\centering
\renewcommand{\arraystretch}{0.5}
\caption{Pseudo-Code of Class-wise Contrastive Loss.} 
\label{Alg: CCL}
\begin{lstlisting}[language={Python}]
# x: feature
# y: label
# n: number of classes in the label
# f_e: feature from encoder
# f_d: feature from decoder
# l: 

def _class_feature(x, y):
   onehot_label = one_hot(y, num_class=n)
   onehot_label = rearrange(onehot_label, 
                  `b n h w' -> `b h w n')
   f = matmul(x, onehot_label)
   f = MLP(LayNorm(f))
   f = Linear(f.view(-1, f.shape[2]))
   return f
def _contrastive_loss(f_e, f_d):
   q = _class_feature(f_e)
   k = _class_fearure(f_d)
   q = normalize(q, dim=1)
   k = normalize(k, dim=1)
   logits =matmul(q, k.transpose(0, 1))
   labels = arange(logits.shape[0])
   loss = CrossEntropyLoss(logits, labels)
   return loss
\end{lstlisting}
\end{algorithm}

\noindent \textbf{Class-wise Contrastive Loss.}
We believe that the final layer of the Mini-Decoder exhibits superior class feature representation capability. Consequently, we employ the deepest layer features from the Encoder and the features from this specific layer to compute the Class-wise Contrastive Loss (CCL). This incentivizes features within the same class to share more related information, while reducing the sharing of information between different classes. The ultimate goal is to maximize intra-class similarity and minimize inter-class similarity. The design of this loss primarily comprises two components: the extraction of class-specific features and the computation of the contrastive loss.

\noindent \textbf{Extraction of Class-Specific Features:}
We perform matrix multiplication on the one-hot encoded label matrix and the feature matrix to obtain the matrix representation corresponding to each class, denoted as $ \left \{ \boldsymbol{F}_{\rm{En}}, \boldsymbol{F}_{\rm{De}} \right \} $, 
where $\text{En}$ and $\text{Dn}$ signify the features of the Encoder and the Decoder.

\noindent \textbf{Computation of the Contrastive Loss:}
We utilize a multilayer perceptron layer to extract mutual information between the two class features followed by spatial alignment, resulting in feature mapping $ \left \{ {{\boldsymbol{\hat{F}}}}_{\rm{En}}, {{\boldsymbol{\hat{F}}}}_{\rm{De}} \right \} $. Subsequently, we apply the InfoNCE Loss~\cite{oord2018representation} for contrastive learning to constrain the class feature relationships within the same branch: 
\begin{equation}
    \mathcal{L}_{\text{CC}} =
    \frac{\boldsymbol{\hat{F}}_{\text{En}}^{(p,c)}\cdot \boldsymbol{\hat{F}}^{(p,c) \top}_{\text{De}}}
    {\sum^\mathrm{P}_{p=1} \sum^\mathrm{C}_{c=1} \boldsymbol{\hat{F}}_\text{En}^{(p,c)}\cdot \boldsymbol{\hat{F}}^{(p,c) \top}_\text{De}},
\end{equation}
where the meanings of $p$ and $c$ are consistent with the previous descriptions. To facilitate a deeper understanding of this loss design, we present the pseudocode in \cref{Alg: CCL}.

\noindent \textbf{Overall Learning Objective.}
In practice, we integrate the standard cross-entropy loss, in addition to the two previously mentioned losses, to train the deep network. The standard cross-entropy loss is applied to the classification layer, while the Hybrid Disparity-enhancing Loss is employed on both the classification layer and the feature representation layer. We assemble the Cross-Entropy Loss and our proposed Hybrid Disparity-enhancing Loss, $\left \{ {\mathcal{L}}_{\text{CE}}, {\mathcal{L}}_{\text{HD}} \right \}$, as our overall learning objective:
\begin{equation}
    \mathcal{L} = {\mathcal{L}}_{\text{CE}} + \lambda {\mathcal{L}}_{\text{HD}},
\end{equation}
where $\lambda$ is a balancing weight, set as 1 in our experiments.

\section{Experiment}

\begin{table*}[t]
\caption{\textbf{Quantitaive results} on WHU-OHS Dataset. Average accuracy and class average F1-score (CF1) are proposed}
\label{Tab: Complete WHU_OHS Quantitative}
\centering
\setlength{\tabcolsep}{2.5mm}{
\begin{tabular}{c|ccccccc >{\columncolor{lightblue}}c}
\toprule[1.2pt]
Method & SSFCN~\cite{xu2019beyond} & A2S2K~\cite{roy2020attention} & SSDGL~\cite{zhu2021spectral} & 3D-CNN~\cite{chen2016deep} & CLSJE~\cite{yu2022cross}  & FreeNet~\cite{zheng2020fpga} & 3D-FCN~\cite{zou2020spectral} & CSCN (Ours)\\ 
\midrule
1  & 0.720 & 0.798 & 0.825 & 0.701 & 0.847 & 0.713 & 0.843 & 0.845 \\ 
2  & 0.696 & 0.786 & 0.827 & 0.680 & 0.843 & 0.679 & 0.844 & 0.841 \\ 
3  & 0.696 & 0.751 & 0.769 & 0.718 & 0.787 & 0.747 & 0.794 & 0.798 \\ 
4  & 0.408 & 0.481 & 0.524 & 0.494 & 0.579 & 0.551 & 0.585 & 0.582 \\ 
5  & 0.037 & 0.297 & 0.052 & 0.367 & 0.132 & 0.419 & 0.035 & 0.250 \\ 
6  & 0.000 & 0.197 & 0.011 & 0.418 & 0.137 & 0.466 & 0.032 & 0.213 \\ 
7  & 0.547 & 0.662 & 0.678 & 0.430 & 0.701 & 0.561 & 0.673 & 0.696 \\ 
8  & 0.326 & 0.407 & 0.457 & 0.426 & 0.473 & 0.513 & 0.473 & 0.481 \\ 
9  & 0.377 & 0.567 & 0.593 & 0.574 & 0.606 & 0.599 & 0.604 & 0.622 \\ 
10 & 0.614 & 0.740 & 0.759 & 0.693 & 0.790 & 0.687 & 0.777 & 0.809 \\ 
11 & 0.915 & 0.918 & 0.926 & 0.894 & 0.950 & 0.893 & 0.937 & 0.952 \\ 
12 & 0.548 & 0.603 & 0.657 & 0.739 & 0.726 & 0.774 & 0.698 & 0.724 \\ 
13 & 0.480 & 0.201 & 0.588 & 0.666 & 0.588 & 0.678 & 0.536 & 0.763 \\ 
14 & 0.526 & 0.545 & 0.682 & 0.550 & 0.708 & 0.866 & 0.686 & 0.712 \\ 
15 & 0.659 & 0.814 & 0.870 & 0.784 & 0.870 & 0.792 & 0.878 & 0.873 \\ 
16 & 0.340 & 0.519 & 0.684 & 0.616 & 0.706 & 0.792 & 0.692 & 0.702 \\ 
17 & 0.278 & 0.312 & 0.463 & 0.624 & 0.529 & 0.600 & 0.520 & 0.538 \\ 
18 & 0.791 & 0.841 & 0.844 & 0.803 & 0.881 & 0.831 & 0.877 & 0.891 \\ 
19 & 0.653 & 0.792 & 0.818 & 0.502 & 0.835 & 0.732 & 0.834 & 0.808 \\ 
20 & 0.538 & 0.605 & 0.598 & 0.615 & 0.531 & 0.662 & 0.462 & 0.726 \\
21 & 0.355 & 0.448 & 0.581 & 0.677 & 0.662 & 0.752 & 0.597 & 0.703 \\
22 & 0.000 & 0.567 & 0.254 & 0.787 & 0.308 & 0.778 & 0.256 & 0.541 \\
23 & 0.674 & 0.777 & 0.825 & 0.728 & 0.840 & 0.779 & 0.832 & 0.845 \\
24 & 0.918 & 0.967 & 0.973 & 0.965 & 0.977 & 0.948 & 0.976 & 0.985 \\
\midrule
CF1 & 0.504 & 0.629 & 0.636 & 0.644 & 0.644 & 0.667 & 0.683 & 0.704 \\
\bottomrule[1.2pt]
\end{tabular}
}
\end{table*}

To assess the effectiveness of our model, we conduct a comparative analysis against state-of-the-art models using the large-scale land-cover HSI classification dataset, WHU-OHS dataset~\cite{WHU-OHS}. Furthermore, in order to validate the transferability and robustness of our model, we conduct comprehensive experiments across eight reputable benchmark datasets. We select random subsets of labeled samples to underscore the model's exceptional performance.
Our comparative analysis includes both established deep-learning methods and recent innovations, covering CNN-based and Transformer-based strategies. In HSI classification, methods are typically categorized as patch-based or image-based. Although our method is of the latter category, comparisons have been made with both types of methods.
All the experiments are performed on NVIDIA 3090 GPU with 24GB RAM.

\subsection{Results on WHU-OHS Dataset}

\begin{figure*}[t]
  \centering
  \includegraphics[width=0.95\linewidth]{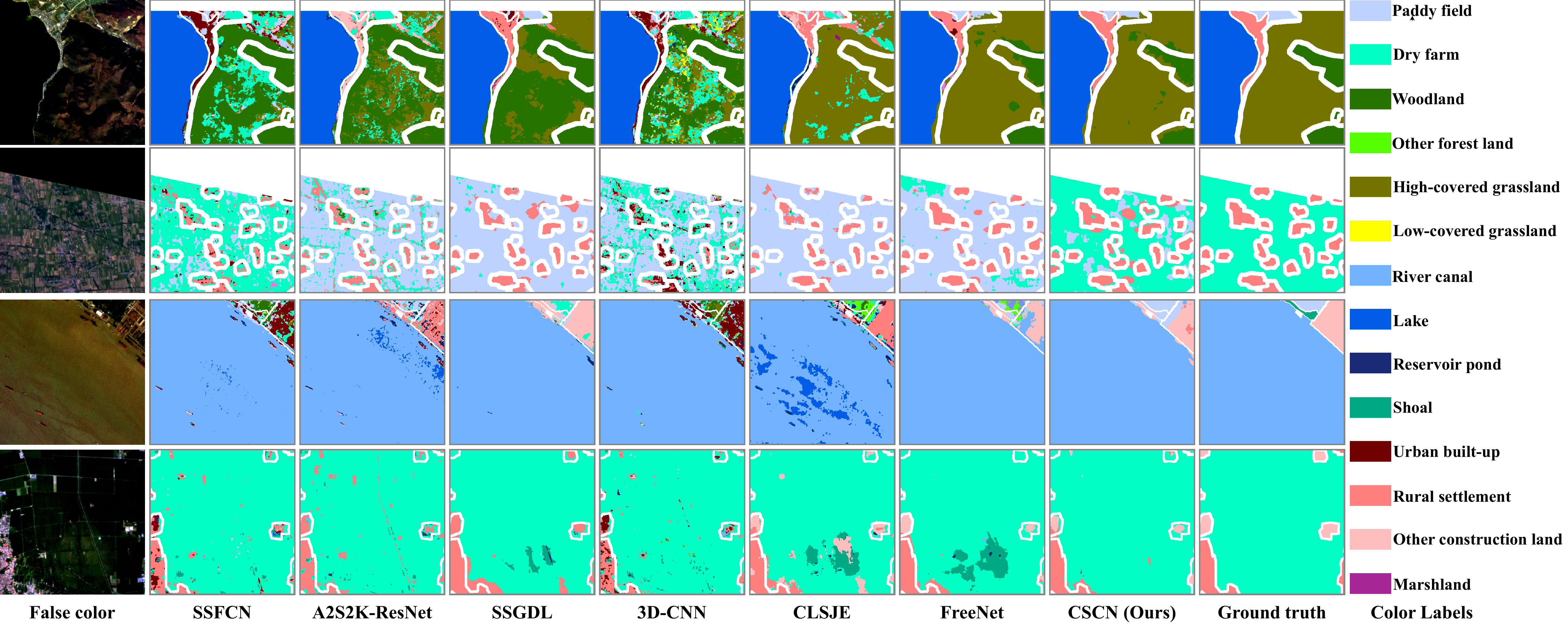}
  \caption{\textbf{Visualization results} on WHU-OHS dataset. In cases where the similarity in magnitude spectral distribution leads to feature confusion, our method benefits from the assistance of the derivative spectrum, achieving more accurate classification.}
  \label{Fig: WHU-OHS Qualitative}
\end{figure*}

\noindent\textbf{Data and Setups.}
The WHU-OHS dataset is an extensive land-cover HSI classification dataset, consisting of 7795 Orbita Hyperspectral Satellite (OHS) image patches from over 40 locations in China, featuring 24 land-cover types. Each image patch is of dimension $512 \times 512$ pixels with 32 spectral channels. We train the model for 50 epochs using the Adam optimizer~\cite{kingma2014adam} with a learning rate of 0.0001. We report the F1 score for each class and the class-average F1 score (CF1).

\noindent\textbf{Quantitative Evaluation.}
As depicted in~\cref{Tab: Complete WHU_OHS Quantitative}, the proposed method, CSCN, achieves a CF1 score of \textbf{0.704}, showing an improvement over the suboptimal result by \textbf{2.1\%}. Meanwhile, our method consistently achieves performance improvements across most classes.
Specifically, previous methods suffered from insufficient utilization of spectral information and faced challenges with spectral mixing.
A2S2K-ResNet and FreeNet address the extraction of spatial and spectral features, yet they lack joint attention, leading to suboptimal utilization of spectral and spatial information. 
In contrast, 3DCNN and 3DFCN excel in jointly extracting spectral and spatial features, effectively leveraging information from both domains. SSDGL and CLSJE achieve joint learning of spectral and spatial information, mitigating spatial information loss while capturing spectral-related features. However, challenges such as information entanglement and feature loss persist.
Introducing spectral derivative features enhances spectral details and reduces overlap. The proposed fusion module effectively utilizes attention mechanisms to fully exploit the complementary information between magnitude and derivative features, providing the network with more discriminative features.

\noindent\textbf{Qualitative Evaluation.}
The visual results obtained from the methods on the WHU-OHS dataset are illustrated in \cref{Fig: WHU-OHS Qualitative}. Particularly, our method demonstrates a more comprehensive representation of land cover structure and class boundaries that closely resemble real images. Furthermore, our results are less noisy and uniform inside the image. These accomplishments can be attributed to the introduction of spectral derivative features and the design of loss functions that explore the internal relationships within the dual branches. These aspects assist the network in fully utilizing both the spectral macrostructure and microdetails, thereby highlighting the effectiveness and superiority of our approach in HSI classification.

\subsection{Results on Other Six Benchmark Datasets}

\noindent\textbf{Data and Setups.}
Using the IP dataset as a case study, our model processes the complete image, employing a masking technique to segment training and testing datasets. Initially, we pinpoint non-background pixel indices. Then, 10\% of these indices per class are chosen at random for the training set, with the rest allocated to the testing set. This method yields training and testing sets that match the image's dimensions, with labels designated at select indices.
For image-based methods, we adhere to the described procedure. Patch-based methods use a consistent $ 16 \times 16 $ pixel size across all approaches to ensure fairness. We allocate 5\% of the samples for Loukia training. For the PU, HanChuan, LongKou, and Dioni datasets, the label split is 1\%, while the Augsburg and Berlin datasets use a 10\% ratio.

For image-based methods, we employ the SGD optimizer~\cite{wilson2017marginal} with a learning rate of 0.001, training for 600 epochs. Patch-based methods, due to their increased sample size, undergo training for 200 epochs with all other settings consistent.
Performance is gauged using four established metrics: overall accuracy (OA), average accuracy (AA), and Kappa coefficient (Kappa). Experiments are repeated five times to counter estimation bias, with outcomes presented as the ``mean accuracy".

\begin{table*}[t]
\centering
\caption{\textbf{Quantitaive results} on IP and PU Datasets, with training and testing sets split ratio in parentheses.}
\label{Tab: Benchmark Quantitative 1}
\setlength{\tabcolsep}{5.6mm}{
\begin{tabular}{rl|ccc|ccc}
\toprule[1.2pt]
& \multirow{2}{*}{Method} & \multicolumn{3}{c|}{IP (10\%)} & \multicolumn{3}{c}{PU (1\%)} \\ 
\cmidrule{3-8} 
 & & OA (\%) $\uparrow$ & AA (\%) $\uparrow$ & $\kappa$ $\uparrow$ & OA (\%) $\uparrow$ & AA (\%) $\uparrow$ & $\kappa$ $\uparrow$ \\ 
\midrule
\multirow{7}{*}{\rotatebox{90}{Patch-based}} & SSRN~\cite{zhong2017spectral} & 98.18 & 90.83 & 97.93 & 92.43 & 89.90 & 89.85 \\
& DBDA~\cite{li2020classification} & 98.20 & 96.31 & 97.95 & 96.06 & 94.71 & 94.76 \\
& A2S2K-ResNet~\cite{roy2020attention} & 98.63 & 97.72 & 98.12 & 95.90 & 95.10 & 94.56 \\
& SSTN~\cite{zhong2021spectral} & 98.69 & 97.87 & 98.51 & 97.47 & 94.87 & 96.65 \\
& SSFTT~\cite{sun2022spectral} & 98.75 & 97.33 & 98.58 & 97.68 & 95.35 & 96.92 \\
& morphFormer~\cite{roy2023spectral} & 98.98 & 97.85 & 98.84 & 95.90 & 93.87 & 94.56 \\
& GSCViT~\cite{zhao2024hyperspectral} & 98.54 & 95.68 & 98.34 & 97.58 & 96.01 & 96.79 \\
\midrule
\multirow{6}{*}{\rotatebox{90}{Image-based}} & ENL-FCN~\cite{shen2020efficient} & 98.36 & 87.17 & 98.13 & 98.56 & 97.28 & 98.08 \\
& FreeNet~\cite{zheng2020fpga} & 98.40 & 97.92 & 98.17 & 97.85 & 96.55 & 97.15 \\
& SSDGL~\cite{zhu2021spectral} & 98.85 & 98.82 & 98.69 & 98.31 & 97.44 & 97.76 \\
& CLSJE~\cite{yu2022cross} & 98.76 & 98.84 & 98.59 & 97.08 & 96.07 & 96.14 \\
& SAHRNet~\cite{tu2024multi} & 98.95 & 95.73 & 98.81 & 97.38 & 95.39 & 96.50 \\
\cmidrule{2-8} 
\rowcolor{lightblue} \cellcolor[RGB]{255,255,255} & CSCN~(Ours) & \textbf{99.19} & \textbf{98.85} & \textbf{99.08} & \textbf{98.82} & \textbf{98.12} & \textbf{98.44} \\
\bottomrule[1.2pt]
\end{tabular}
}
\end{table*}

\begin{table*}[t]
\centering
\caption{\textbf{Quantitaive results} on WHU-Hi-HanChuan and WHU-Hi-LongKou Datasets, with training and testing sets split ratio in parentheses.}
\label{Tab: Benchmark Quantitative 2}
\setlength{\tabcolsep}{5.6mm}{
\begin{tabular}{rl|ccc|ccc}
\toprule[1.2pt]
& \multirow{2}{*}{Method} & \multicolumn{3}{c|}{WHU-Hi-HanChuan (1\%)} & \multicolumn{3}{c}{WHU-Hi-LongKou (1\%)} \\ 
\cmidrule{3-8} 
 & & OA (\%) $\uparrow$ & AA (\%) $\uparrow$ & $\kappa$ $\uparrow$ & OA (\%) $\uparrow$ & AA (\%) $\uparrow$ & $\kappa$ $\uparrow$ \\ 
\midrule
\multirow{7}{*}{\rotatebox{90}{Patch-based}} & SSRN~\cite{zhong2017spectral} & 96.93 & 94.36 & 96.41 & 99.38 & 98.00 & 99.18 \\
& DBDA~\cite{li2020classification} & 98.26 & 97.21 & 97.97 & 99.19 & 97.43 & 98.93 \\
& A2S2K-ResNet~\cite{roy2020attention} & 95.76 & 92.42 & 94.93 & 99.48 & 98.49 & 99.31 \\
& SSTN~\cite{zhong2021spectral} & 93.95 & 91.04 & 92.90 & 98.80 & 97.47 & 98.43 \\
& SSFTT~\cite{sun2022spectral} & 96.64 & 93.77 & 96.07 & 99.46 & 98.54 & 99.29 \\
& morphFormer~\cite{roy2023spectral} & 97.45 & 95.85 & 97.01 & 99.35 & 98.25 & 99.15 \\
& GSCViT~\cite{zhao2024hyperspectral} & 97.35 & 95.65 & 96.90 & 99.39 & 98.65 & 99.20 \\
\midrule
\multirow{6}{*}{\rotatebox{90}{Image-based}} & ENL-FCN~\cite{shen2020efficient} & 95.51 & 83.47 & 94.74 & 99.48 & 98.26 & 99.32 \\
& FreeNet~\cite{zheng2020fpga} & 96.89 & 86.89 & 96.36 & 99.33 & 97.82 & 99.12 \\
& SSDGL~\cite{zhu2021spectral} & 98.48 & 95.57 & 98.22 & 99.51 & 98.09 & 99.35 \\
& CLSJE~\cite{yu2022cross} & 98.53 & 95.91 & 98.28 & 99.14 & 97.07 & 98.87 \\
& SAHRNet~\cite{tu2024multi} & 98.40 & 95.48 & 98.12 & 99.49 & 98.73 & 99.33 \\
\cmidrule{2-8} 
\rowcolor{lightblue} \cellcolor[RGB]{255,255,255} & CSCN~(Ours) & \textbf{98.75} & \textbf{96.87} & \textbf{98.54} & \textbf{99.60} & \textbf{98.92} & \textbf{99.48} \\
\bottomrule[1.2pt]
\end{tabular}
}
\end{table*}

\begin{table*}[t]
\centering
\caption{\textbf{Quantitaive results} on HyRANK-Dioni and HyRANK-Loukia Datasets, with training and testing sets split ratio in parentheses.}
\label{Tab: Benchmark Quantitative 3}
\setlength{\tabcolsep}{5.6mm}{
\begin{tabular}{rl|ccc|ccc}
\toprule[1.2pt]
& \multirow{2}{*}{Method} & \multicolumn{3}{c|}{HyRANK-Dioni (1\%)} & \multicolumn{3}{c}{HyRANK-Loukiau (5\%)} \\ 
\cmidrule{3-8} 
 & & OA (\%) $\uparrow$ & AA (\%) $\uparrow$ & $\kappa$ $\uparrow$ & OA (\%) $\uparrow$ & AA (\%) $\uparrow$ & $\kappa$ $\uparrow$ \\ 
\midrule
\multirow{7}{*}{\rotatebox{90}{Patch-based}} & SSRN~\cite{zhong2017spectral} & 88.27 & 87.37 & 85.52 & 84.90 & 81.55 & 81.69 \\
& DBDA~\cite{li2020classification} & 88.67 & 83.15 & 85.92 & 82.19 & 79.92 & 78.39 \\
& A2S2K-ResNet~\cite{roy2020attention} & 87.58 & 83.36 & 84.59 & 85.64 & 82.46 & 82.55 \\
& SSTN~\cite{zhong2021spectral} & 88.69 & 83.70 & 86.02 & 85.52 & 81.16 & 82.41 \\
& SSFTT~\cite{sun2022spectral} & 90.18 & 87.85 & 87.81 & 85.77 & 82.54 & 82.73 \\
& morphFormer~\cite{roy2023spectral} & 90.23 & 87.96 & 87.82 & 84.51 & 82.63 & 81.16 \\
& GSCViT~\cite{zhao2024hyperspectral} & 90.52 & 85.20 & 88.24 & 85.54 & 82.08 & 82.43 \\
\midrule
\multirow{6}{*}{\rotatebox{90}{Image-based}} & ENL-FCN~\cite{shen2020efficient} & 90.43 & 82.20 & 88.10 & 88.36 & 82.89 & 85.83 \\
& FreeNet~\cite{zheng2020fpga} & 90.74 & 73.91 & 88.47 & 89.82 & 80.52 & 87.60 \\
& SSDGL~\cite{zhu2021spectral} & 91.12 & 77.21 & 88.92 & 90.06 & 82.16 & 87.91 \\
& CLSJE~\cite{yu2022cross} & 89.27 & 78.68 & 84.90 & 89.27 & 78.68 & 86.95 \\
& SAHRNet~\cite{tu2024multi} & 90.73 & 76.05 & 88.43 & 88.49 & 72.92 & 85.77 \\
\cmidrule{2-8} 
\rowcolor{lightblue} \cellcolor[RGB]{255,255,255} & CSCN~(Ours) & \textbf{91.83} & \textbf{88.74} & \textbf{89.86} & \textbf{91.36} & \textbf{86.28} & \textbf{89.44} \\
\bottomrule[1.2pt]
\end{tabular}
}
\end{table*}

\begin{table*}[t]
\centering
\caption{\textbf{Quantitaive results} on Augsburg and Berlin Datasets, with training and testing sets split ratio in parentheses.}
\label{Tab: Benchmark Quantitative 4}
\setlength{\tabcolsep}{5mm}{
\renewcommand\arraystretch{1.2}
\begin{tabular}{rl|ccc|ccc}
\toprule[1.2pt]
& \multirow{2}{*}{Method} & \multicolumn{3}{c|}{Augsburg (10\%)} & \multicolumn{3}{c}{Berlin (10\%)} \\ 
\cmidrule{3-8} 
 & & OA (\%) $\uparrow$ & AA (\%) $\uparrow$ & $\kappa$ $\uparrow$ & OA (\%) $\uparrow$ & AA (\%) $\uparrow$ & $\kappa$ $\uparrow$ \\ 
\midrule
\multirow{6}{*}{\rotatebox{90}{Image-based}} & ENL-FCN~\cite{shen2020efficient} & 72.73 & 46.85 & 67.29 & 82.44 & 60.51 & 78.22 \\
& FreeNet~\cite{zheng2020fpga} & 81.96 & 69.84 & 78.53 & 83.98 & 67.34 & 80.15 \\
& SSDGL~\cite{zhu2021spectral} & 81.86 & 69.59 & 78.12 & 84.40 & 68.00 & 80.76 \\
& CLSJE~\cite{yu2022cross} & 82.05 & 75.09 & 78.66 & 85.90 & 71.65 & 82.64 \\
& SAHRNet~\cite{tu2024multi} & 82.11 & 72.10 & 78.77 & 86.24 & 76.97 & 83.14 \\
\cmidrule{2-8} 
\rowcolor{lightblue} \cellcolor[RGB]{255,255,255} & CSCN~(Ours) & \textbf{84.12} & \textbf{78.16} & \textbf{81.19} & \textbf{87.08} & \textbf{80.29} & \textbf{84.20} \\
\bottomrule[1.2pt]
\end{tabular}
}
\end{table*}

\begin{figure}[h]
  \centering
  \includegraphics[width=0.90\linewidth]{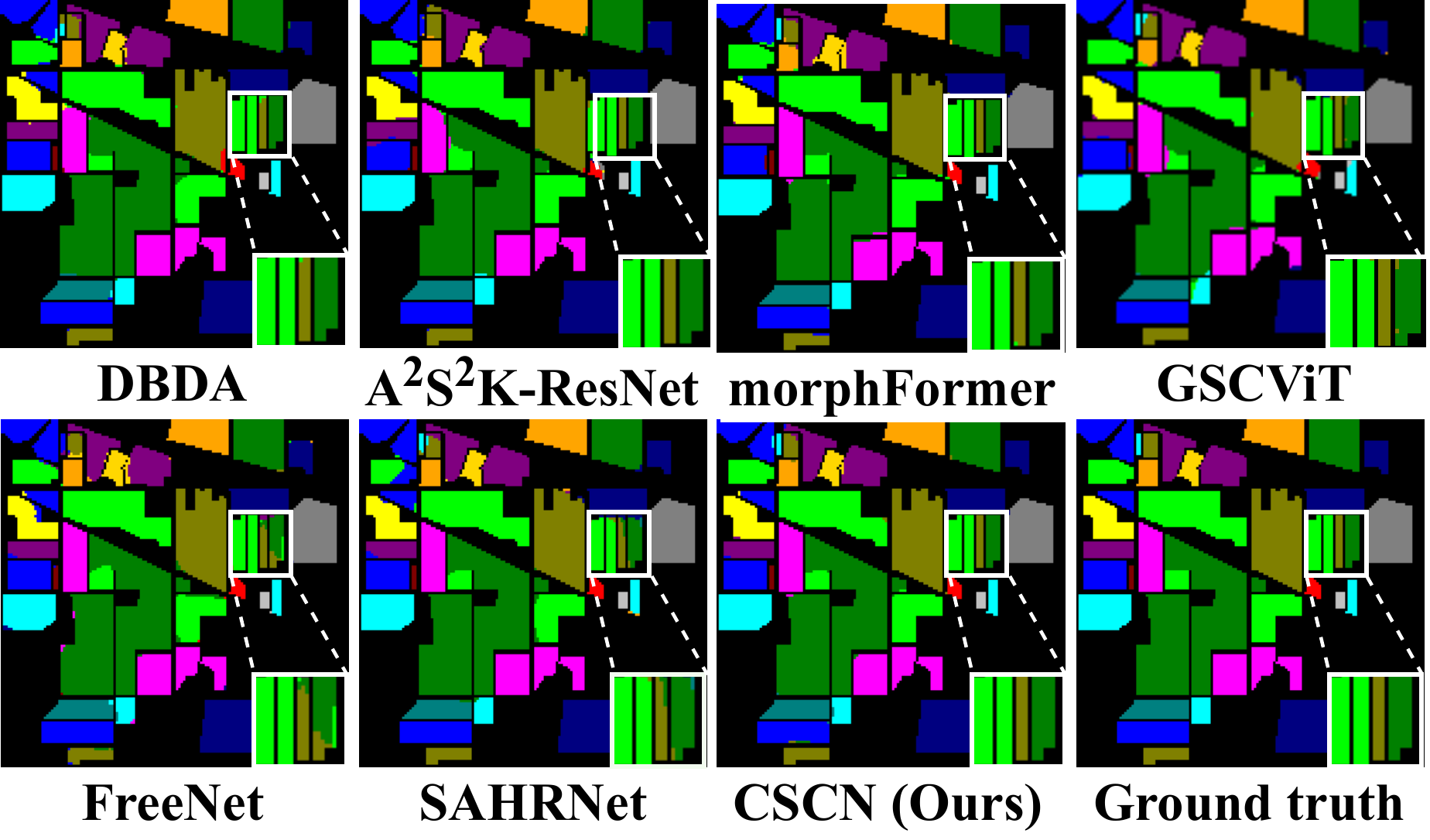}
  \caption{\textbf{Visualization results} on IP dataset. For a detailed inspection, we zoom in on a specific section of the map.}
  \label{Fig: IP Qualitative}
\end{figure}

\begin{figure}[!h]
  \centering
  \includegraphics[width=0.90\linewidth]{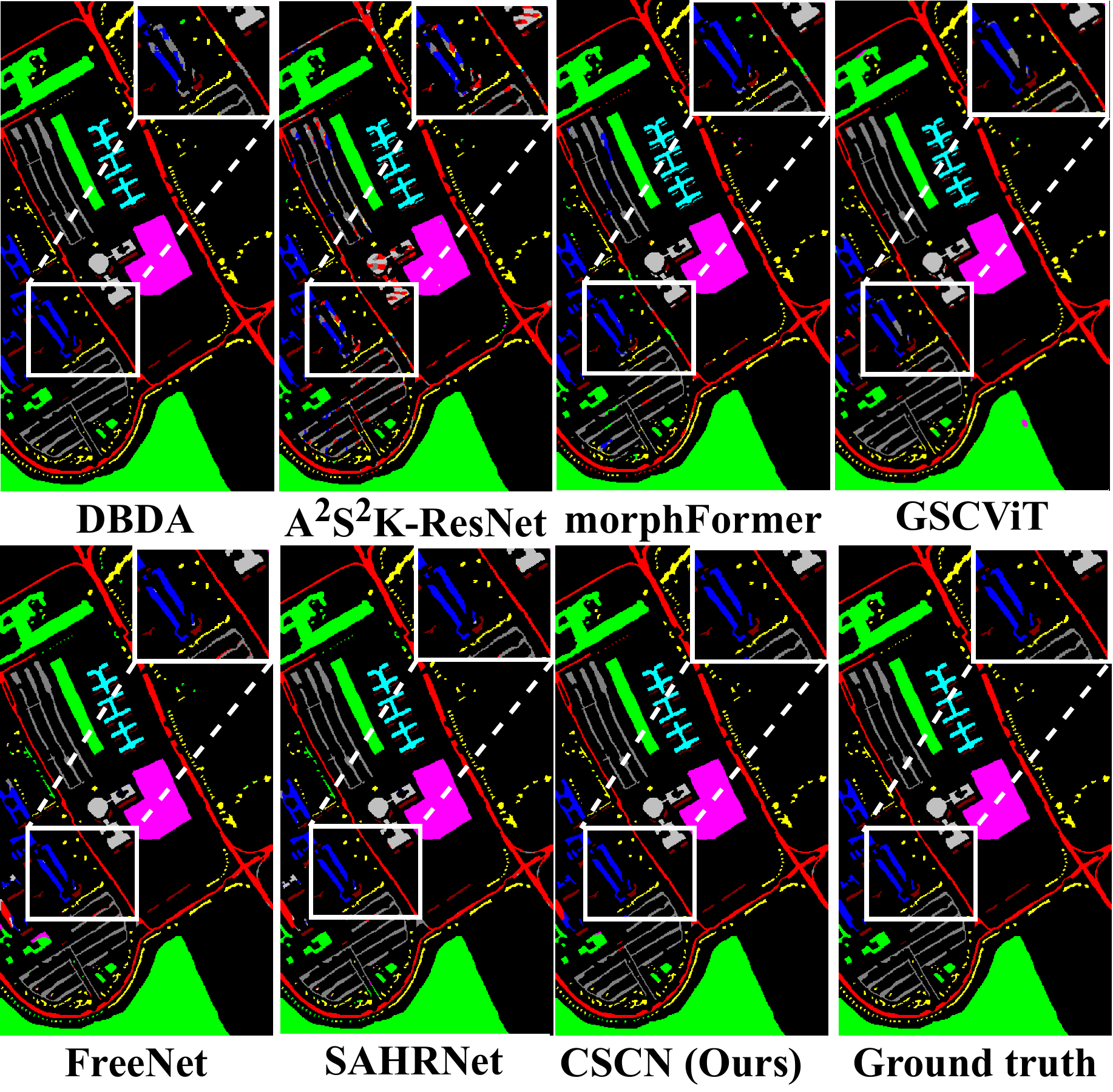}
  \caption{\textbf{Visualization results} on PU dataset. For a detailed inspection, we zoom in on a specific section of the map.}
  \label{Fig: PU Qualitative}
\end{figure}

\begin{figure}[!h]
  \centering
  \includegraphics[width=0.85\linewidth]{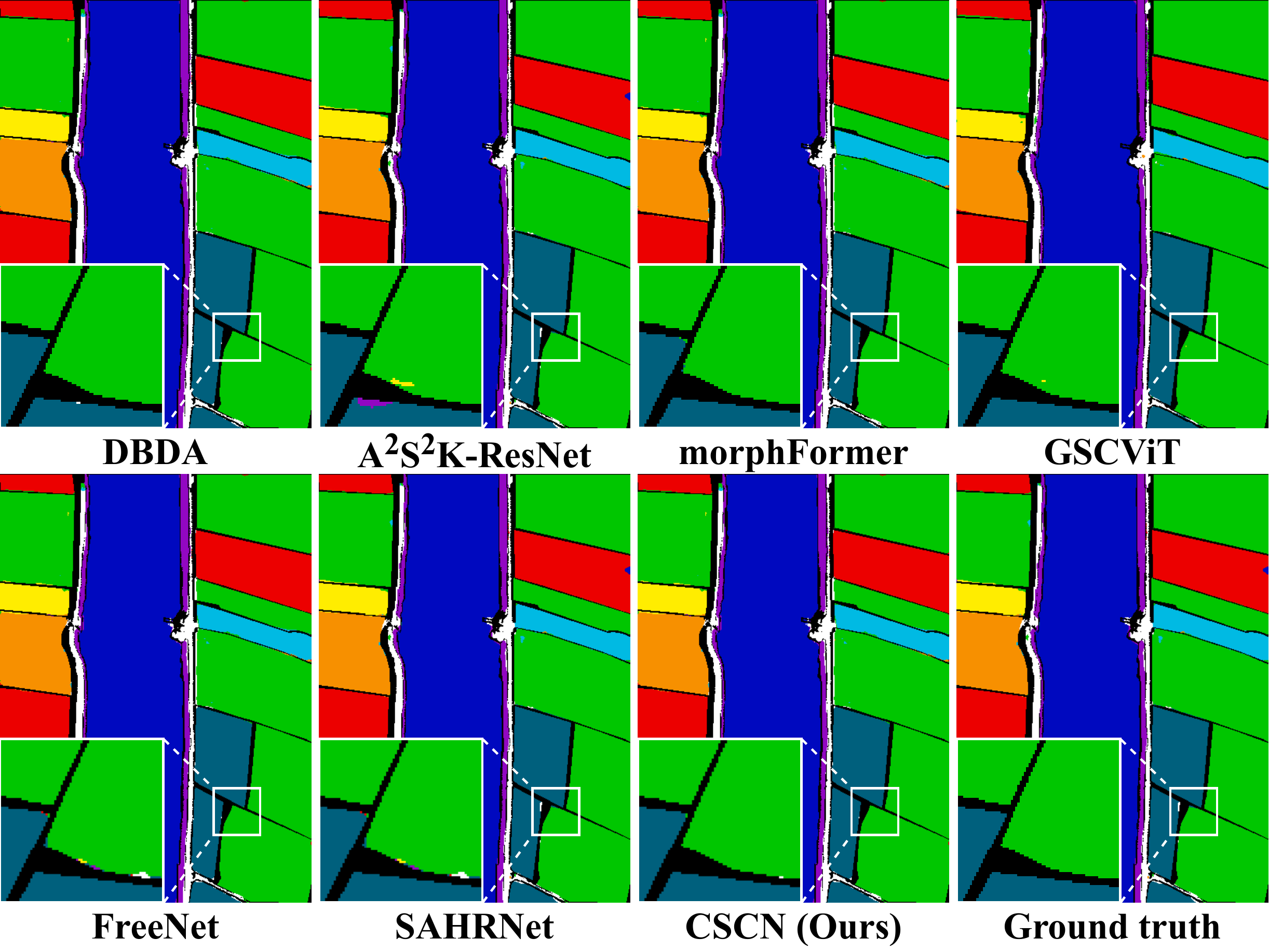}
  \caption{\textbf{Visualization results} on WHU-Hi LongKou dataset. For a detailed inspection, we zoom in on a specific section of the map.}
  \label{Fig: LongKou Qualitative}
\end{figure}

\begin{figure}[!h]
  \centering
  \includegraphics[width=0.95\linewidth]{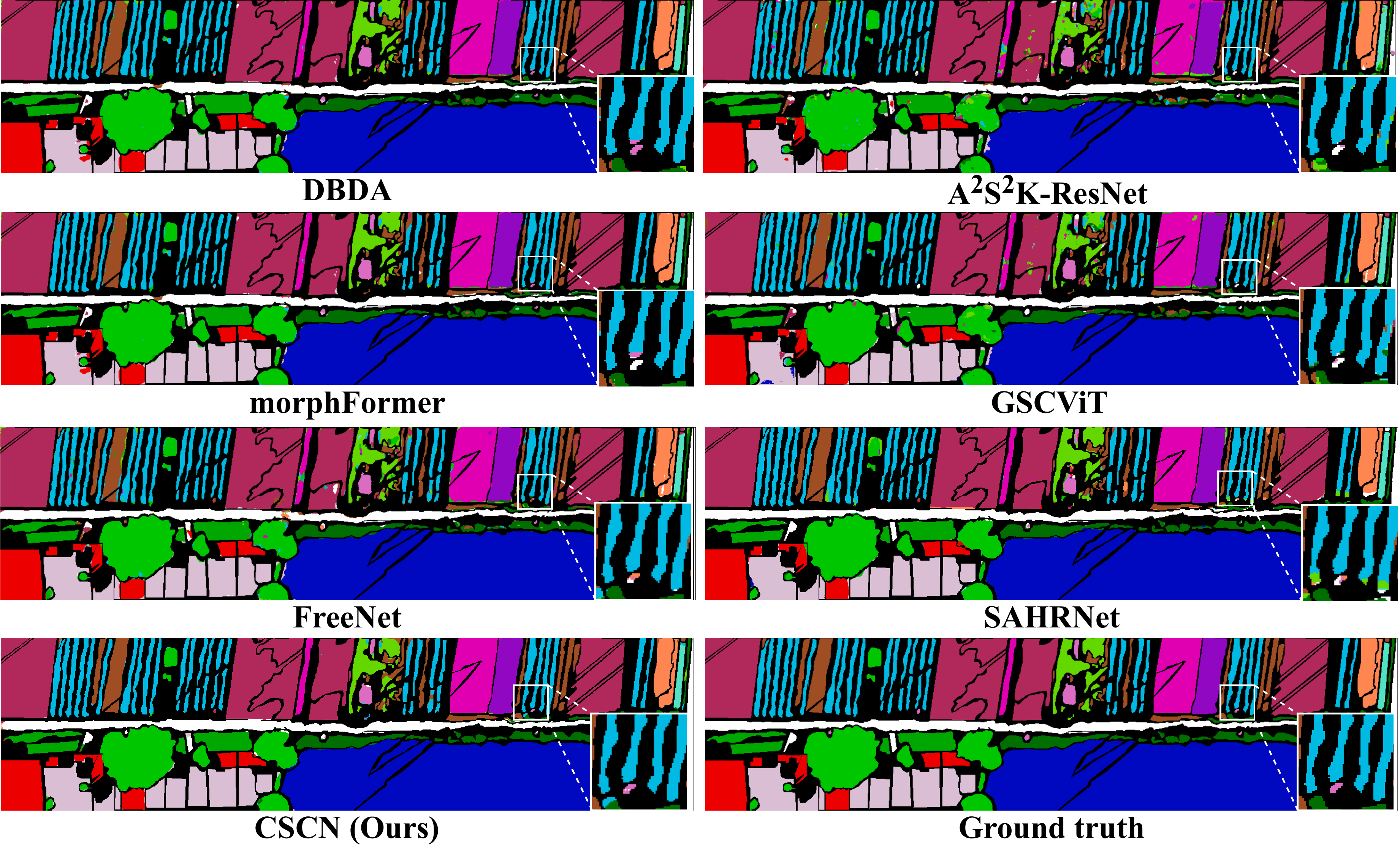}
  \caption{\textbf{Visualization results} on WHU-Hi HanChuan dataset. For a detailed inspection, we zoom in on a specific section of the map.}
  \label{Fig: HanChuan Qualitative}
\end{figure}

\begin{figure}[!h]
  \centering
  \includegraphics[width=0.95\linewidth]{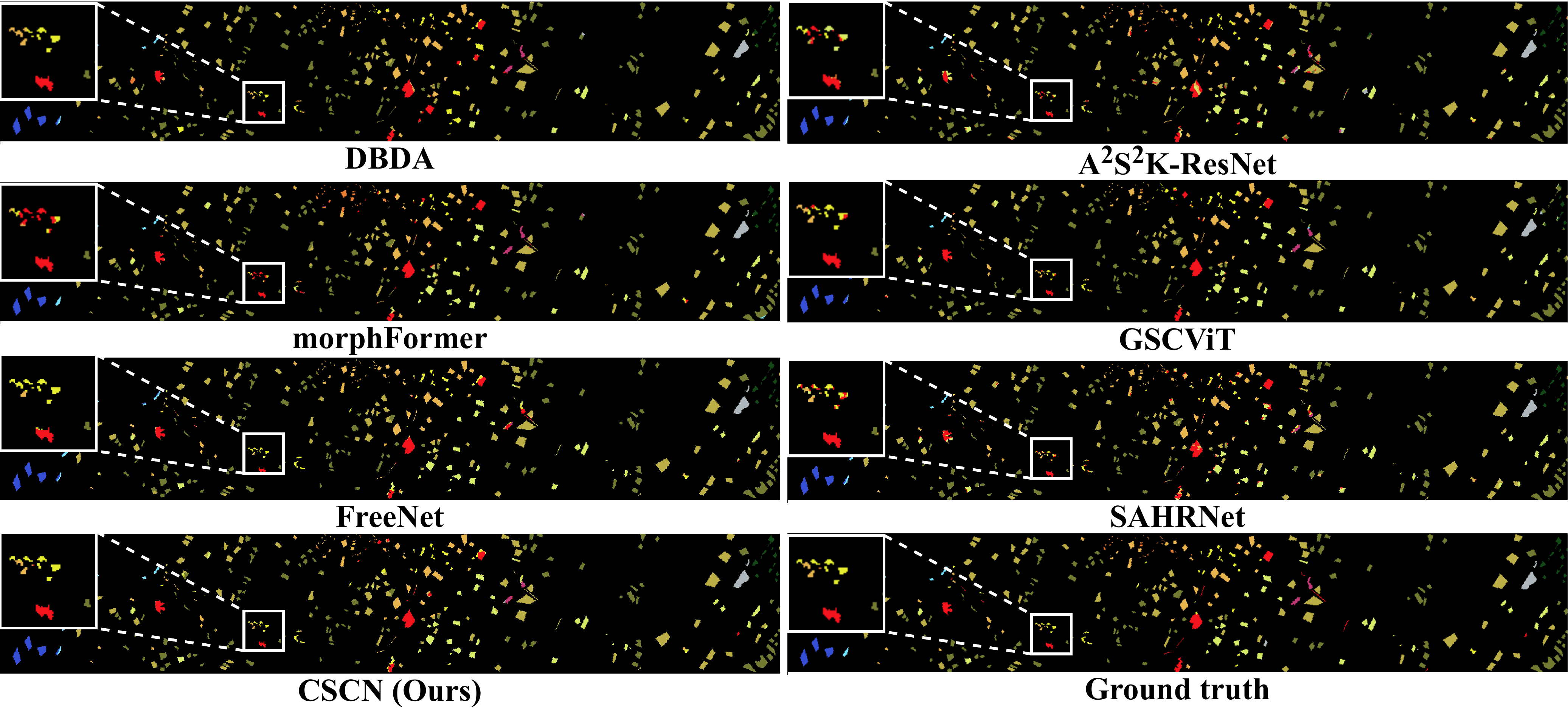}
  \caption{\textbf{Visualization results} on HyRANK-Dioni dataset. For a detailed inspection, we zoom in on a specific section of the map.}
  \label{Fig: Dioni Qualitative}
\end{figure}

\begin{figure}[!h]
  \centering
  \includegraphics[width=0.95\linewidth]{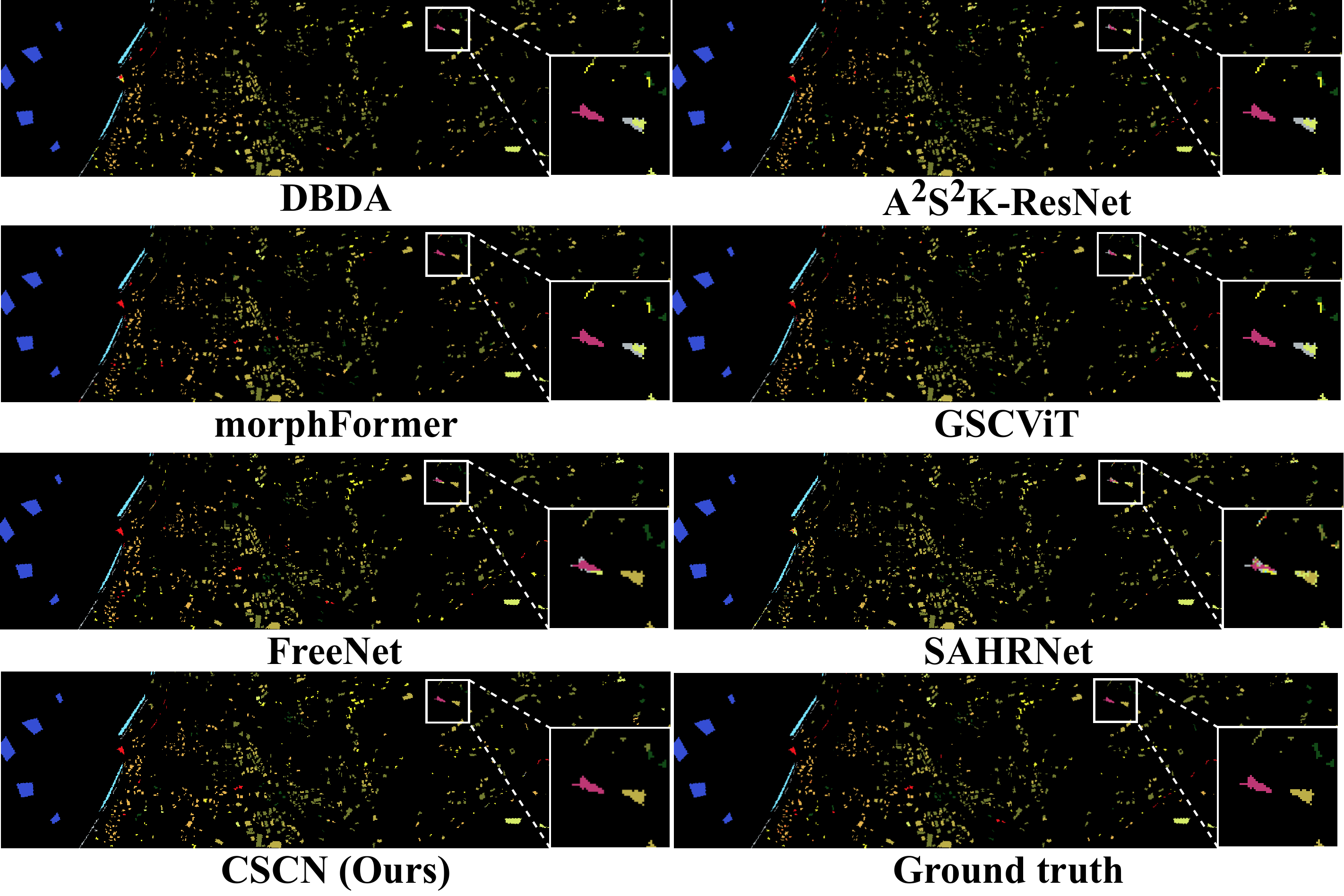}
  \caption{\textbf{Visualization results} on HyRANK-Loukia dataset. For a detailed inspection, we zoom in on a specific section of the map.}
  \label{Fig: Loukia Qualitative}
\end{figure}

\noindent\textbf{Quantitative Results.}
As shown in \cref{Tab: Benchmark Quantitative 1}, \cref{Tab: Benchmark Quantitative 2}, and \cref{Tab: Benchmark Quantitative 3}, the method we propose demonstrated superior performance across six datasets in terms of OA, AA, and Kappa.
The notable improvement highlights the exceptional feature extraction ability of our method, particularly evident in cases with imbalanced samples.
To be more precise, for IP, PU, WHU-Hi-HanChuan, and WHU-Hi-LongKou, where performance has already approached saturation, we achieve OA improvements of 0.3\%, 0.26\%, 0.22\%, and 0.09\%. For the HyRANK-Dioni and HyRANK-Loukia datasets, we realize OA increases of 0.71\% and 1.3\%, AA enhancements of 0.78\% and 3.39\%, and Kappa gains of 0.94 and 1.53.
This success is primarily attributed to the incorporation of the spectral derivative branch, which facilitates the comprehensive utilization of spectral information. Concurrently, the introduction of the loss function enhances the diversity of feature representations between the two branches while encouraging inter-class separation, thereby achieving a more comprehensive extraction of discriminative features and higher classification accuracy.

\noindent\textbf{Qualitative Evaluation.}
We present the visual comparison between other results and our method in \cref{Fig: IP Qualitative}, \cref{Fig: PU Qualitative}, \cref{Fig: HanChuan Qualitative}, \cref{Fig: LongKou Qualitative}, \cref{Fig: Dioni Qualitative}, \cref{Fig: Loukia Qualitative}. For aesthetic purposes, we have selected a subset of methods to present the visualization results. For a detailed inspection of the classification result maps, we zoom in on a specific section of the map, encompassing multiple categories. In these zoomed-in sections, our method showcases a distinct advantage. It demonstrates the ability to discern between land cover types that are in close spatial proximity but belong to different classes, delineating precise boundaries between classes.

\subsection{Extended Experiments on Image-based Methods}
To further validate the effectiveness of our method, we conduct experiments using image-based methods on two datasets with a substantial number of annotated samples. The Augsburg dataset~\cite{hong2023cross} comprises $886 \times 1360$ pixels, with a total of 224 spectral bands and 1,199,777 annotated samples. Additionally, the Berlin dataset~\cite{hong2023cross} consists of $2465 \times 811$ pixels, also with 224 spectral bands, and includes 1,614,315 annotated samples. The extensive sample annotations in these datasets provide a more robust validation of the effectiveness and scientific soundness of our method.
As shown in \cref{Tab: Benchmark Quantitative 4}, our method demonstrates significant improvements over other methods across both datasets. In the Augsburg dataset, the OA increased by 2.01\%, AA by 3.07\%, and the Kappa by 2.42\%. In the Berlin dataset, the OA improved by 0.84\%, AA by 3.32\%, and kappa by 1.06\%.
We attribute these performance enhancements to the extraction and integration of spectral derivative features and the introduction of Class-wise Contrastive Loss. The incorporation of spectral derivative features allows the network to capture more comprehensive spectral information, thereby facilitating the extraction of discriminative features. The significant improvement in AA across both datasets underscores the effectiveness of our method in mitigating class confusion caused by spectral overlap and similar distributions. This is further supported by the Class-wise Contrastive Loss, which bolsters the network's capacity to represent and extract distinguishable features between classes.

\begin{table}[t]
    \centering
    \caption{Discussion on Spectral Degradation.}
    \centering
    \label{Tab: spectral degradation}
    \setlength{\tabcolsep}{9.0mm}
    {
    \renewcommand\arraystretch{1.2}
    \begin{tabular}{c|c}
    \toprule[1.2pt]
    Method & CF1 $\uparrow$ \\
    \midrule
    CSCN (Ours) & \textbf{0.704} \\
    CSCN (Ours) + Spectral Degradation & 0.698 \\
    \bottomrule[1.2pt]
    \end{tabular}
    }
\end{table}

\subsection{Extended Experiments on Spectral Noise.}
Inspired by \cite{hong2019mixing}, spectral data are often subject to disturbances and noise. We simulate spectral degradation by adding a mixture of Gaussian noise, salt-and-pepper noise, and strip noise to the original data to explore the impact of spectral variation on the model. The comparative results, as shown in \cref{Tab: spectral degradation}, indicate a 0.6\% reduction in performance. Despite this reduction, when compared with the performance of other methods on the undegraded dataset in \cref{Tab: Complete WHU_OHS Quantitative}, our method retains a certain advantage. This can be ascribed to the introduction of spectral derivative features while amplifying subtle details, inevitably amplifying noise as well. Furthermore, the adaptive fusion between the dual-branch features enhances the model's ability to extract discriminative features, thereby improving the model's robustness and resistance to noise.

\begin{table}[!t]
    \centering
    \caption{Ablation studies on key components.}
    \centering
    \label{Tab: Ablation}
    \setlength{\tabcolsep}{3.5mm}{
    \renewcommand\arraystretch{1.68}
    \begin{tabular}{cccc|cc}
    \toprule[1.2pt]
    Baseline & Dual-Encoder & CPFM & HDLoss  & CF1 $\uparrow$ \\
    \midrule
    \textcolor{red}{\ding{51}} & \textcolor{green}{\ding{55}} & \textcolor{green}{\ding{55}} & \textcolor{green}{\ding{55}} & 0.665 \\
    \textcolor{red}{\ding{51}} & \textcolor{red}{\ding{51}} & \textcolor{green}{\ding{55}} & \textcolor{green}{\ding{55}} & 0.681 \\
    \textcolor{red}{\ding{51}} & \textcolor{red}{\ding{51}} & \textcolor{red}{\ding{51}} & \textcolor{green}{\ding{55}} & 0.697 \\
    \textcolor{red}{\ding{51}} & \textcolor{red}{\ding{51}} & \textcolor{red}{\ding{51}} & \textcolor{red}{\ding{51}} & \textbf{0.704} \\
    \bottomrule[1.2pt]
    \end{tabular}
    }
\end{table}

\begin{figure}[t]
  \centering
  \includegraphics[width=1.0\linewidth]{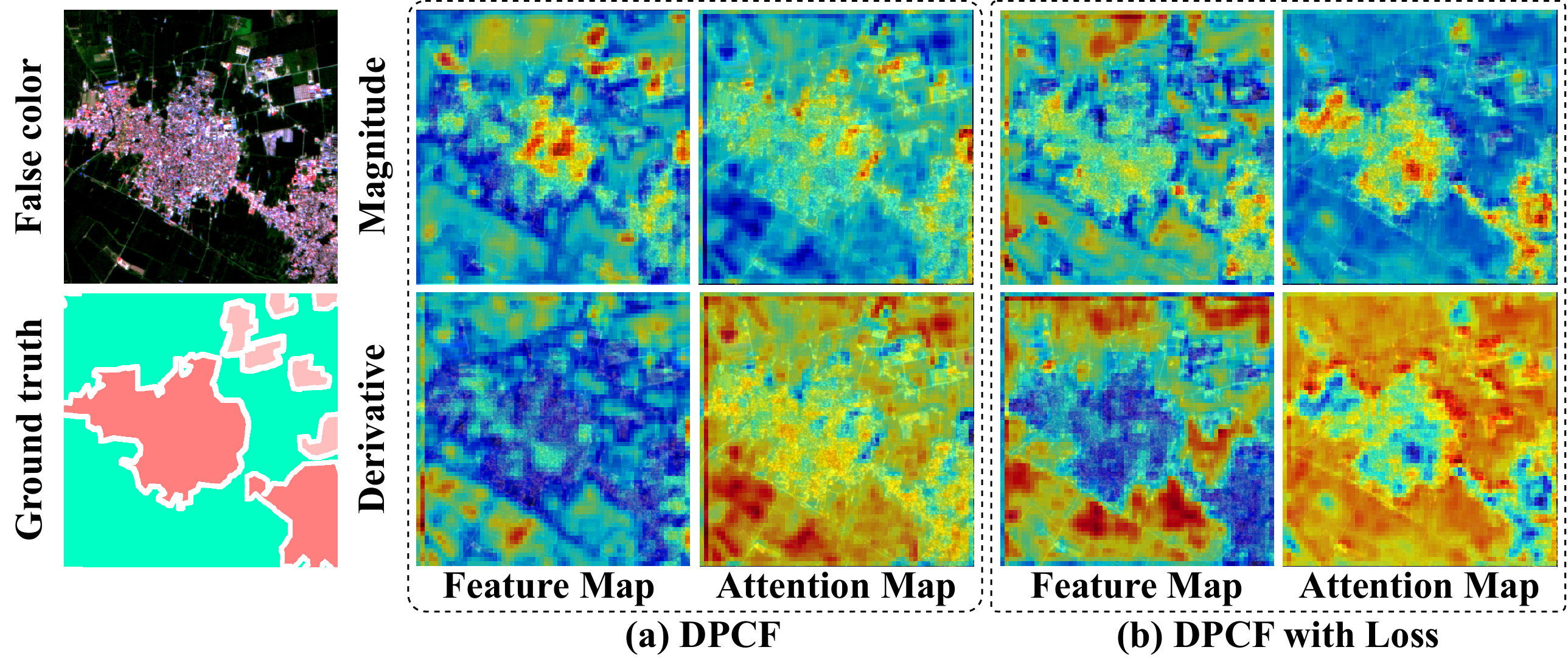}
  \caption{Visualization of Feature Maps and Attention Maps. \textbf{(a)} Context-aware Point-wise Feature Fusion Module enhances feature aggregation capacity. \textbf{(b)} The Hybrid Disparity-enhancing Loss amplifies the focus on complementary information.}
  \label{Fig: Attention Map}
\end{figure}

\begin{figure*}[t]
  \centering
  \includegraphics[width=0.90\linewidth]{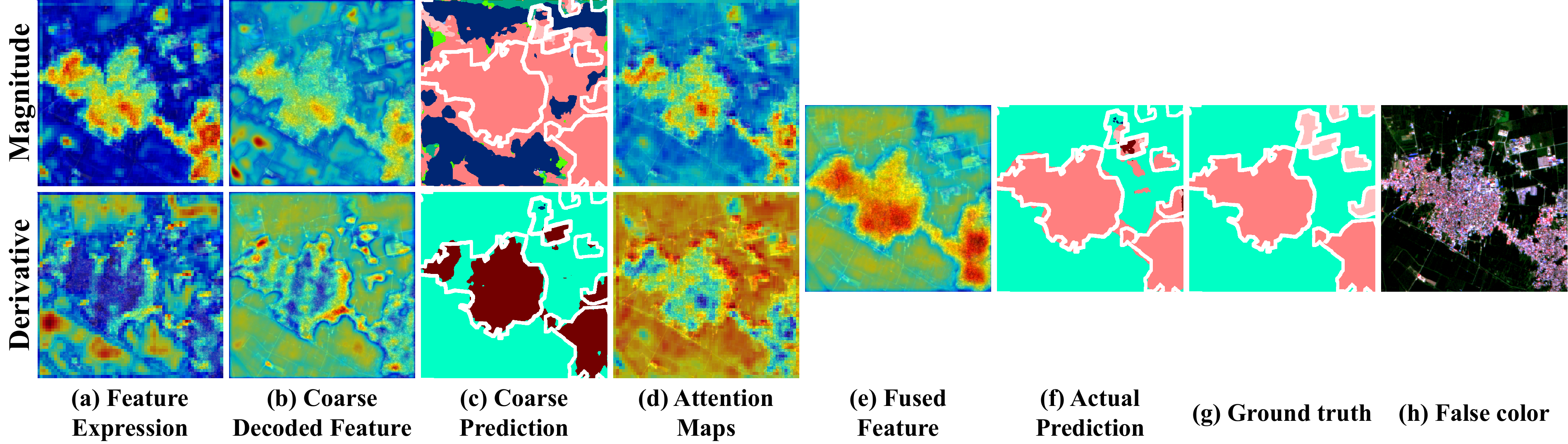}
  \caption{Visual results on feature map, attention map, and prediction. \textbf{(a)} represent the individual feature expressions of the dual-encoder. Employing the mini-decoder detailed in \cref{Loss}, we obtain coarse decoded features \textbf{(b)} and predicted results \textbf{(c)}. Utilizing the fusion module described in \cref{Fusion}, we obtain the point-wise weight map shown in \textbf{(d)}. Additionally, the fused features and the actual prediction results are showcased in \textbf{(e)} and \textbf{(f)}. The ground truth is depicted in \textbf{(g)}.}
  \label{Fig: Complete Method visual}
\end{figure*}

\subsection{Ablation Studies}
We conduct an ablation study on the WHU-OHS dataset, meanwhile employing a simple single-branch Encoder-Decoder architecture as our baseline.

\noindent\textbf{Effect of Dual-Encoder.}
To determine the impact of spectral magnitude and derivative features on model learning, we conduct experiments with each of these features as separate inputs to a single-stream network, as well as combined inputs in the Dual-Encoder network. As shown in \cref{Tab: Ablation}, upon incorporating a dual encoder that inputs both spectral magnitude and derivative features, the performance improvement is 2.4\%. It explicitly demonstrates a significant improvement in model performance with the combined input. Additionally, it indicates that the introduction of spectral derivative features provides the model with more spectral details, significantly enhancing spectral information utilization and providing valuable complementary information.

\noindent\textbf{Effect of Content-adaptive Point-wise Fusion Module.}
We investigate the effectiveness of our fusion module. \cref{Tab: Ablation} demonstrates a 1.6\% increase in the F1 score when this module is added to the model. 
Furthermore, as illustrated in the visual comparison between the feature maps and attention maps of the two branches, shown in \cref{Fig: Attention Map}~\red{(a)}. It can be observed that the incorporation of the module achieves adaptive fusion based on the content of each point, thereby obtaining more discriminative features.

\noindent\textbf{Effect of Hybrid Disparity-enhancing Loss.}
\cref{Tab: Ablation} demonstrates the impact of our loss function on network performance within the context of the fusion module. The results suggest that the introduction of loss functions contributes to performance improvement. As shown in \cref{Fig: Attention Map}~\red{(b)}, the loss function appears to better enhance the differentiation information in the features of the two branches.

\begin{table}[tp]
    \centering
    \caption{Impact of the Order of Derivative.}
    \label{Tab: Ablation Derivate}
    \setlength{\tabcolsep}{2.0mm}
    {
    \renewcommand\arraystretch{1.2}
    \begin{tabular}{cccc|c}
    \toprule[1.2pt]
    First-order & Second-order & Dual-Encoder &  Fusion \& Loss & CF1 $\uparrow$ \\
    \midrule
    \textcolor{red}{\ding{51}}   & \textcolor{green}{\ding{55}} &  \textcolor{green}{\ding{55}} & \textcolor{green}{\ding{55}} & 0.665 \\
    \textcolor{green}{\ding{55}} & \textcolor{red}{\ding{51}}   & \textcolor{green}{\ding{55}} & \textcolor{green}{\ding{55}} & 0.659 \\
    \textcolor{red}{\ding{51}}   & \textcolor{green}{\ding{55}} & \textcolor{red}{\ding{51}}   & \textcolor{green}{\ding{55}} & 0.681 \\
    \textcolor{green}{\ding{55}} & \textcolor{red}{\ding{51}}   & \textcolor{red}{\ding{51}}   & \textcolor{green}{\ding{55}} & 0.680 \\
    \textcolor{red}{\ding{51}}   & \textcolor{green}{\ding{55}} & \textcolor{red}{\ding{51}}   & \textcolor{red}{\ding{51}}   & 0.704 \\
    \textcolor{green}{\ding{55}} & \textcolor{red}{\ding{51}}   & \textcolor{red}{\ding{51}}   & \textcolor{red}{\ding{51}}   & 0.689 \\
    \bottomrule[1.2pt]
    \end{tabular}
    }
\end{table}

\begin{table}[tp]
    \centering
    \caption{Impact of Input Format.}
    \centering
    \label{Tab: Ablation Input Format}
    \setlength{\tabcolsep}{8.0mm}
    {
    \renewcommand\arraystretch{1.2}
    \begin{tabular}{c|c}
    \toprule[1.2pt]
    Input Format & CF1 $\uparrow$ \\
    \midrule
    Concatenation & 0.671 \\
    Dual-Encoder \textit{with} shared parameters & 0.674 \\
    \rowcolor{lightblue} Dual-Encoder \textit{without} shared parameters & \textbf{0.681} \\
    \bottomrule[1.2pt]
    \end{tabular}
    }
\end{table}

\begin{table*}[t]
    \begin{minipage}[t]{0.33\textwidth}
        \centering
        \caption{Loss Weight $\lambda$.}
        \centering
        \label{Tab: Loss Weight}
        \setlength{\tabcolsep}{11.5mm}
        {
        \renewcommand\arraystretch{1.2}
        \begin{tabular}{c|c}
        \toprule[1.2pt]
        $\lambda$ & CF1 $\uparrow$ \\
        \midrule
        0.5 & 0.686\\
        \rowcolor{lightblue} \textbf{1} & \textbf{0.704}\\
        2   & 0.679\\
        3   & 0.678\\
        \bottomrule[1.2pt]
        \end{tabular}
        }
    \end{minipage}
    \begin{minipage}[t]{0.33\textwidth}
        \centering
        \caption{Convolution Block Numbers $\mathrm{N}$.}
        \centering
        \label{Tab: Convolution Block Numbers}
        \setlength{\tabcolsep}{11.5mm}
        {
        \renewcommand\arraystretch{1.2}
        \begin{tabular}{c|c}
        \toprule[1.2pt]
        $\mathrm{N}$ & CF1 $\uparrow$ \\
        \midrule
        3 & 0.677 \\
        \rowcolor{lightblue} \textbf{4} & \textbf{0.704} \\
        5 & 0.689 \\
        6 & 0.683 \\
        \bottomrule[1.2pt]
        \end{tabular}
        }
    \end{minipage}
    \begin{minipage}[t]{0.33\textwidth}
        \centering
        \caption{Feature dimensions $\mathrm{C_f}$.}
        \centering
        \label{Tab: Feature dimensions}
        \setlength{\tabcolsep}{11.5mm}
        {
        \renewcommand\arraystretch{1.2}
        \begin{tabular}{c|c}
        \toprule[1.2pt]
        $\mathrm{C_f}$ & CF1 $\uparrow$ \\
        \midrule
        64  & 0.680 \\
        \rowcolor{lightblue} 128 & \textbf{0.704} \\
        192 & 0.696 \\
        256 & 0.701 \\
        \bottomrule[1.2pt]
        \end{tabular}
        }
    \end{minipage}
\end{table*}

\noindent\textbf{Visualization of Module Functionality.}
In addition to quantitative metric results and qualitative visualization outcomes, we aim to intuitively demonstrate the effectiveness and mechanism of action of our proposed module through the visualization of feature maps.
As shown in \cref{Fig: Complete Method visual}~\red{(a)} and \cref{Fig: Complete Method visual}~\red{(b)}, the features extracted by individual branches exhibit certain limitations. This is evident from the rough predictions in \cref{Fig: Complete Method visual}~\red{(c)}, indicating distinct discriminatory advantages for different classes between the two branches, showcasing strong complementarity. Leveraging our proposed fusion module, we explore the inherent relationship between the two branches and extract the adaptive pixel-wise fusion in \cref{Fig: Complete Method visual}~\red{(d)}, represented by the weight map. This realization of adaptive dual-branch fusion produces the fused features \cref{Fig: Complete Method visual}~\red{(e)} and the final prediction results \cref{Fig: Complete Method visual}~\red{(f)}. These results underscore the comprehensive utilization of complementary information by the fusion module. Furthermore, the comparison between prediction results \cref{Fig: Complete Method visual}~\red{(c)} and \cref{Fig: Complete Method visual}~\red{(f)} also highlights the significance of our proposed loss, enhancing the divergence between the two branches and providing rich complementary information for the fusion module.

\noindent\textbf{Impact of the Order of Derivative.}
As mentioned earlier, commonly employed spectral derivative analyses include first-order and second-order derivatives. \cref{Tab: Ablation Derivate} illustrates the impact of the derivative order on model learning. We conduct experiments under various component conditions, replacing the first-order derivatives with second-order derivatives. Under the baseline, performance decreased by 0.6\%, under the dual encoder, it decreased by 0.1\%, and with the feature fusion module and loss function, the performance decline was 1.5\%. The results indicate that the first-order derivative while providing additional detailed features and supplementary information, reduces the disturbance of spectral noise in the extraction of spectral features.

\noindent\textbf{Input Format of Magnitude and Derivative Spectrum.}
In our network, we employ a symmetric dual-encoder structure to achieve the complementary utilization of two pieces of information. Consequently, investigate the possibility of integrating and inputting two sets of spectral information into the network in a concatenated manner. As shown in \cref{Tab: Ablation Input Format}, the accuracy of the network with both features concatenated as input is significantly lower than that of the designed Dual-Encoder structure. We attribute this result to the indiscriminate feature extraction process in the concatenated feature input, which, despite enriching the feature information, neglects the extraction of unique information in spectral difference features. Consequently, it fails to fully utilize the complementary information between spectral magnitude and derivative features. To further validate our approach, we conducted a comparative experiment by introducing a dual-encoder structure with shared parameters between branches. The experimental results indicate that independently extracting features is essential, promoting the diversified extraction and utilization of features and complementary information.

\noindent\textbf{Impact of Hybrid Disparity-enhancing Loss  Weight.}
\cref{Tab: Loss Weight} illustrates the impact of balancing weights on the loss. The results indicate that variations in weights may affect the delicate balance between the dual-encoder fusion classification loss and the loss of maintaining the differences between the branches, leading to a decrease in model performance. In our experiments, the model achieved the best performance when $\lambda$ was set to 1.

\noindent\textbf{Impact of Model Configuration.}
We primarily investigate the number of convolution blocks mentioned in \cref{encoder}, the feature dimensions discussed in \cref{feature embedding}, and the selection of convolution kernel sizes within the convolution blocks.

\noindent \textbf{(i) Impact of Convolution Block Numbers.}
We investigate the impact of the number of convolution blocks. As shown in \cref{Tab: Convolution Block Numbers}, we vary the number of convolution blocks $\mathrm{N}$, mentioned in \cref{encoder}, and analyzed the resulting CF1 score. The results indicate that when the number of convolution blocks is 6, CF1 decreases to 0.683, representing a 2.1\% performance drop compared to using 4 convolution blocks. This result demonstrates that having too many convolution blocks can have a negative impact on the feature extraction process of the network. Despite increasing the parameter count and deepening the network, positive effects are not observed.

\noindent \textbf{(ii) Impact of Feature dimensions.}
We study the impact of the hyperparameter $\mathrm{C_f}$ settings mentioned in \cref{Fusion} when all components are included. \cref{Tab: Feature dimensions} indicates that the performance is optimal when this parameter is set to 128.

\noindent \textbf{(iii) Impact of Convolution Kernel Size.}
Based on the previous discussion, using four convolutional blocks yielded the best performance. Building on this, we examine the impact of convolution kernel size on model performance. When all four convolutional blocks use 3x3 kernels, the model achieved the best performance with a CF1 score of 0.704. Using 5x5 kernels for all blocks results in a CF1 score of 0.673. A mixed usage of 3x3 and 5x5 kernels yielded a performance of 0.683 or 0.680. This indicates that the smaller 3x3 kernel size is more effective in our framework for capturing essential features.

\section{Conclusion}
This work introduces the Content-driven Spectrum Complementary Network for HSI classification. 
The Dual-Encoder structure combines spectral magnitude and derivative features, extracting rich complementary information. 
The Content-adaptive Point-wise Fusion module enables the adaptive fusion of complementary information. 
The proposed Hybrid Disparity-enhancing Loss enhances feature representation differentiation, enriches complementary information, and increases inter-class distances. 
Extensive experiments on eight HSI classification datasets validate our method's effectiveness, demonstrating its capability to leverage spectral information. 
Looking ahead, the foundational model SpectralGPT~\cite{hong2024spectralgpt}, specifically designed for spectral remote sensing data, offers a transformative solution for the efficient and comprehensive utilization of spectral data. Building on this, we can integrate its custom feature learning modules, tailored for spectral data, to create pre-trained models specifically designed for applications in spectral remote sensing.

\bibliographystyle{IEEEtran}
\bibliography{references}

\vfill
\end{document}